\documentclass[letterpaper]{article} 
\usepackage{aaai23}  
\usepackage{times}  
\usepackage{helvet}  
\usepackage{courier}  
\usepackage[hyphens]{url}  
\usepackage{graphicx} 
\usepackage{natbib}  
\usepackage{caption} 
\usepackage{algorithm}
\usepackage{algorithmic}

\usepackage{newfloat}
\usepackage{listings}
\DeclareCaptionStyle{ruled}{labelfont=normalfont,labelsep=colon,strut=off} 
\floatstyle{ruled}
\newfloat{listing}{tb}{lst}{}
\floatname{listing}{Listing}
\title{Boundary Graph Neural Networks for 3D Simulations}

\author {
    Andreas Mayr\textsuperscript{\rm 1,$\dag$},
    Sebastian Lehner\textsuperscript{\rm 1},
    Arno Mayrhofer\textsuperscript{\rm 2},
    Christoph Kloss\textsuperscript{\rm 2},\\
    Sepp Hochreiter\textsuperscript{\rm 1,3},
    Johannes Brandstetter\textsuperscript{\rm 1,$\ddagger$,\thanks{now at Microsoft Research AI4Science}}
}
\affiliations {
    \textsuperscript{\rm 1} ELLIS Unit Linz \& LIT AI Lab, Johannes Kepler University Linz, Linz, Austria\\
    \textsuperscript{\rm 2} DCS Computing GmbH, Linz, Austria\\
    \textsuperscript{\rm 3} Institute of Advanced Research in Artificial Intelligence (IARAI), Vienna, Austria\\
    \textsuperscript{\rm $\dag$} mayr@ml.jku.at, \textsuperscript{\rm $\ddag$} brandstetter@ml.jku.at
}

\usepackage{tikz-imagelabels}
\usepackage{subcaption}
\usepackage{overpic}
\usepackage{amsmath}
\usepackage{amssymb}
\usepackage{mathtools}
\usepackage{amsthm}
\usepackage{bm}
\usepackage[capitalize]{cleveref}
\usepackage{booktabs}
\usepackage{fp}
\usepackage{makecell}
\usepackage{multirow}
\newcommand\Ba{\bm{a}}
\newcommand\Bb{\bm{b}}

\newcommand\Be{\bm{e}}

\newcommand\Bg{\bm{g}}
\newcommand\Bh{\bm{h}}

\newcommand\Bm{\bm{m}}
\newcommand\Bn{\bm{n}}

\newcommand\Bp{\bm{p}}

\newcommand\Bt{\bm{t}}

\newcommand\Bv{\bm{v}}

\newcommand\Bx{\bm{x}}

\newcommand\BP{\bm{P}}

\newcommand\BX{\bm{X}}

 \newcommand{\dR}{\mathbb{R}}

\newcommand{\rS}{\mathrm{S}}

\newcommand{\cE}{\mathcal{E}} 
\newcommand{\cG}{\mathcal{G}}

 \newcommand{\cV}{\mathcal{V}}

\newcommand\sgn{\mathop{\mathrm{sgn}\,}}

\newcommand{\NRM}[1]{{{\left\| #1\right\|}}} 
\newcommand{\PAR}[1]{{{\left(#1\right)}}} 

\renewcommand{\leq}{\leqslant}

\newcommand{\R}{\mathbb{R}}

\crefname{chapter}{Chap.}{Chaps.}
\crefname{section}{Sect.}{Sects.}
\Crefname{chapter}{Chapter}{Chapters}
\Crefname{section}{Section}{Sections}
\crefname{appendix}{}{}
\Crefname{appendix}{}{}
\crefalias{appendix}{section}

\usepackage{chapterbib}

\usepackage{selectp}

\begin{document}

\maketitle

\begin{abstract}
The abundance of data has given machine learning considerable momentum in natural sciences and engineering, though modeling of physical processes is often difficult. A particularly tough problem is the efficient representation of geometric boundaries. Triangularized geometric boundaries are well understood and ubiquitous in engineering applications. However, it is notoriously difficult to integrate them into machine learning approaches due to their heterogeneity with respect to size and orientation. In this work, we introduce an effective theory to model particle-boundary interactions, which leads to our new Boundary Graph Neural Networks (BGNNs) that dynamically modify graph structures to obey boundary conditions. The new BGNNs are tested on complex 3D granular flow processes of hoppers, rotating drums and mixers, which are all standard components of modern industrial machinery but still have complicated geometry. BGNNs are evaluated in terms of computational efficiency as well as prediction accuracy of particle flows and mixing entropies. BGNNs are able to accurately reproduce 3D granular flows within simulation uncertainties over hundreds of thousands of simulation timesteps. Most notably, in our experiments, particles stay within the geometric objects without using handcrafted conditions or restrictions.
\end{abstract}

\section{Introduction}
The deep learning revolution~\citep{Krizhevsky2012} 
has dramatically changed scientific fields
such as computer vision, natural language processing,
or medical sciences.
More recently, deep learning research has been expanded 
towards physical simulations 
such as fluid dynamics, deformable materials, 
or aerodynamics~\citep{li2018learning, ummenhofer2019, SanchezGonzalez2020, Pfaff2020}. 
Currently, the progress of deep learning for physical simulations is driven by Graph Neural Networks (GNNs)~\citep{Scarselli2008, Defferrard2016, Kipf2017semisupervised}.
GNNs are very effective when modeling interactions between many entities via forward dynamics~\citep{Battaglia2018}, and as such are a strong building block when it comes to the replacement of slower numerical simulations in various engineering disciplines.
We focus on granular flows, which are ubiquitous 
in nature and consequently in industrial processes. 
The accurate simulations of such versatile granular flows 
forms the backbone of designing and improving industrial machinery.
Complex boundaries are present in every-day's machinery such as rotating drums, mixers or hoppers.
In engineering, these complex boundaries are typically modelled
by triangularizations, which are mathematical well founded
and for which efficient construction and simulation tools are available.
Therefore, triangular meshes are 
standard for representing and modelling industrial machinery.

\paragraph{Effective theory.}

In this work, we introduce an effective theory to model 
particle-boundary interactions, from which we derive
a new approach to accurately and effectively model granular flow
processes within triangularized boundary surfaces.
In physics, effective theories allow the description of phenomena within much simpler frameworks
without a significant loss of precision.
The basic idea is to approximate a physical system by factoring out the degrees of freedom 
that are not relevant in the given setting and problem to solve 
(e.g. using Newton's equations instead of the much more complicated Einstein's equations, or, using simple algebraic equations instead of numerically solving differential equations for particle-particle interactions). Other examples are in the fields of
gravitational wave theory
\begin{figure}[h]
\begin{center}
\includegraphics{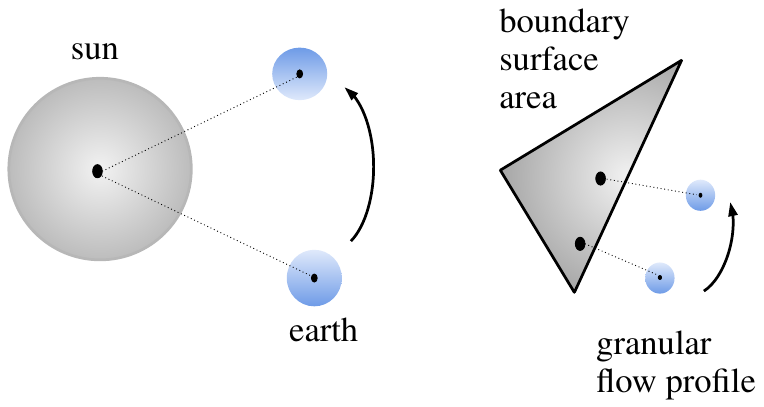}
\end{center}
\caption{Effective theories of gravitational planetary movement (left), and particle-boundary interactions (right). Planetary movement is fully described by Einstein's field equations that relate mass and energy densities to the curvature of spacetime. A much simpler but in most cases sufficient description is to apply Newton's law of gravity to representative point masses.
Black arrows indicate progress in time. Analogously, the interactions of granular flow particles and boundary surface areas is modeled by an effective two-point interaction. \label{fig:effective_theory}} 
\end{figure}
\citep{Goldberger2006}, particle physics \citep{Leutwyler1994}, hydrodynamics \citep{Endlich2013}, and, even in deep learning theory \citep{Roberts2021}. \Cref{fig:effective_theory} illustrates the effective theory of gravitational forces for planetary movement modeling, 
which motivates the introduction of effective particle-boundary interactions 
in this work.

We introduce Boundary Graph Neural Networks (BGNNs) as an effective
model for complex 3D granular flows.
We test the effectiveness of BGNNs on flow simulations within different triangularized geometries. 
The data for BGNN training is obtained 
by precise but potentially time-consuming simulations.
BGNNs are able to generalize granular flow dynamics over thousands of timesteps 
while potentially being considerably faster than state-of-the-art simulation methods.
The contributions of this paper are:\\[0.1cm]
\textbullet\  We describe particle-surface interactions as an effective theory and introduce Boundary Graph Neural Networks (BGNNs)  
which enable dynamic modifications of graph structures.\\[0.1cm]
\textbullet\  We implement BGNNs for 3D granular flow simulations of hoppers, rotating drums, and mixers as found in industrial machinery.\\[0.1cm]
\textbullet\  We assess the performance of BGNNs via comparison of relevant physical quantities 
between model predictions and simulations.

\section{Background}\label{sec:bck}

\textbf{Graph Neural Networks. }
We consider graphs $\cG = (\cV,\cE)$, 
with nodes $v_i \in \cV$ and edges $e_{ij} \in \cE$, 
where $N$-dimensional node features $\Bp_{v_i} \in \dR^{N}$
are attached to each of the nodes.
We use nearest neighbor graphs, assuming local interactions 
allow us to build arbitrary global dynamics.
Therefore, whether an edge between a pair of nodes $(v_i, v_j)$
is contained in the graph $\cG$ depends on the distance between the nodes:
{\fontsize{8.0pt}{9.6pt}\selectfont
\begin{align}
    e_{ij} \in \cE \iff d(v_i, v_j) \leq r_{\text{cut-off}} ,
    \label{eq:real_edges1}
\end{align}
}
\noindent where the cut-off radius $r_{\text{cut-off}}$ is usually a hyperparameter of the model. 
Edges have $M$-dimensional edge features $\Ba_{ij} \in
\dR^{M}$ attached to each edge $e_{ij}$.
Message passing networks~\citep{Gilmer2017neural} are a specific type of graph neural networks~\citep{Battaglia2018} and usually consist of three different types of layers:
i) node and edge feature embedding layers, ii) the core message passing layers, and iii) read-out layers.
Message passing iteratively updates the embeddings of edges ($\Bm_{ij}$) and nodes ($\Bh_i$), i.e., the embeddings of $\Ba_{ij}$ and $\Bp_{v_i}$, at edge $e_{ij}$ and node $v_i$ via:
{\fontsize{8.0pt}{9.6pt}\selectfont
\begin{align}
  \Bm'_{ij} =\ \phi(\Bh_i, \Bh_j, \Bm_{ij} ), \\
 \Bh'_i = \ \psi\Big(\Bh_i,\square_{{e}_{ij} \in {\cE}} \ \Bm'_{ij} \Big) \label{eq:message}, 
\end{align}
}
\noindent where the aggregation $\square_{{e}_{ij} \in {\cE}}$ at node $v_i$ in \cref{eq:message} is across all nodes that are connected to node $v_i$ via an edge $e_{ij}$. Typically, $\square$ represents a mean or max operation. The learnable functions $\phi$ and $\psi$ are commonly presented by Multilayer Perceptrons (MLPs).
\Cref{eq:message} describes the computation and aggregation of messages, and the subsequent update of node embeddings.
The final node embeddings are used for predictions via read-out layers. It is worth noting, that the general concept of GNNs often needs to be adapted to the actual purpose in mind, as e.g. for molecular modeling \citep{Atz2021, Yang2022, Reiser2022}.

\textbf{Dataset.}
Granular flow simulations are obtained by an Discrete Element Method (DEM)~\citep{Cundall1979} which
is similar to molecular dynamics.
For granular flows, governing equations like
the Navier-Stokes equations for fluid flows \citep{Faccanoni2013} do not exist.
DEM represents the granular media by discrete particles (e.g.\ spheres or polyhedra), 
which interact by exchanging momentum via contact models. 
For granular flow simulation with DEM, we resort to the open-source software LIGGGHTS~\citep[][see  \cref{sec:DEM}]{Kloss2012}.
LIGGGHTS can simulate particle flow for a wide range of materials and 
complex mesh-based wall geometries, therefore is well suited to simulate various industrial processes.
In this work, training, validation, and test data are generated by LIGGGHTS
modeling particle trajectories within different machinery designs.

\textbf{Time transition model.}
Our method is based on \citet{SanchezGonzalez2020}, where we
use the semi-implicit Euler method to numerically integrate the equations of motion 
with model-predicted acceleration.
The time-transition from time $t$ to time $t+1$ is given by $\dot{\Bx}^{t+1} =\dot{\Bx}^{t}+\Delta t\; \ddot{\Bx}^{t}$ and $\Bx^{t+1} =\Bx^{t}+\Delta t\; \dot{\Bx}^{t+1}$, 
where $\Bx$ is the particle location, and $\dot{\Bx}$ the 
particle velocity. 
The time-transition $\Bx^{t+1}$ is calculated from 
the predicted particle acceleration $\ddot{\Bx}^{t}$.

\section{Boundary Graph Neural Networks}
\label{sec:BGNNs}

\textbf{Modeling approach. }
Our goal is to model time transition dynamics of particles in complex geometries 
with GNNs. The focus is on developing a proper representation 
of the triangularized geometries. 
An obvious and straightforward approach is to sample individual points 
from the boundaries
as in \citet{SanchezGonzalez2020} and \citet{ummenhofer2019}.
In our setting we have to sample points from the triangles and then include them 
as non-kinematic particles with fixed positions in the graph.
However, sampling is not feasible for large and complex geometries with
many triangles. Therefore, we resort to an effective theory 
to make boundary representations efficient.

\textbf{Effective theory for particle-surface interactions. }
In order to apply an effective theory to particle-boundary interactions, 
we have to determine the most important interaction properties that should 
be conserved. For this purpose, we will define an effective and dynamic graph,
which changes for every timepoint. The graph has to accurately model: \\[0.1cm]
\textbullet\  \textit{Time awareness}: particle-boundary interactions should be modeled
    for as many timesteps as necessary.
    Particle-boundary interactions are represented in the graph as
    connections between surface areas and the particles in their proximity.
    Thus, particle-boundary interactions have to stay within a predefined cutoff radius 
    as long as possible.\\[0.1cm]
\textbullet\  \textit{Capture of strongest interaction}: similar to Newton's law of gravity,
    we target a point-like representation for both particles and surface areas. 
    Knowingly, the physical interaction strength decreases with increasing distance. 
    Thus, effective particle-boundary interactions should contain 
    the smallest distances 
    between particles and surface areas. 

Given these considerations, we model particle-boundary interactions
by point-like particle-particle interactions, 
where the \textit{virtual particles} representing the boundary surface area 
are placed such that the distance to the \textit{real particles} is minimized.
Consequently, real particles ``see'' different virtual particles 
from the same surface area. 
However, for every granular flow particle, we effectively model  
only one particle-surface interaction. 
We give a roadmap of what follows: 
(i) we introduce an efficient way of calculating shortest distances between 
real particles and triangularized surface areas, and 
(ii) we construct a dynamic graph model which models 
the time transition dynamics.

\paragraph{Calculation of shortest distances.} 
In order to obtain shortest distances between real particles and triangularized surface areas, the squared distance between the particle center and the closest point on the mesh triangles is calculated (adopted from \citet{Eberly1999}).
We outline this in the following.
A location on a triangle $\Bt$ is parameterized 
by two scalar values $u_0$, $u_1 \in \dR$ with 
    $\Bt(u_0,u_1) \ = \ \Bb \ + \ u_0 \ \Be_0 \ + \ u_1 \ \Be_1 \ $,
where $u_0\geq0$, $u_1\geq0$, and $u_0+u_1\leq1$, 
$\Bb$ represents one of the nodes of the triangle, 
and, $\Be_0$ and $\Be_1$ are vectors from $\Bb$ 
towards the other two nodes (see \cref{fig:myTriangle}). 
The minimal  Euclidean squared distance $d$ 
of the point $\Bp$ to the triangle is given by the 
optimization problem:
{\fontsize{8.0pt}{9.6pt}\selectfont
\begin{align}
&d \ = \ \min_{u_0,u_1} \ \  q (u_0,u_1) \ = \ \NRM{\Bt(u_0,u_1) \ - \ \Bp}^2 \label{eq:triPointMin} \\ \nonumber
&\mbox{s.t.} \ \ \ \   u_0\geq0 \ , \ \ u_1\geq0 \ , \ \ u_0+u_1\leq1 \ . 
\end{align}
}
The minimizing arguments $u_0'$ and $u_1'$ 
parameterize the closest point $\Bt(u_0',u_1')$ 
of the triangle to the point $\Bp$. The algorithmic computation of this minimization problem is more involved and comprises seven cases, that need to be distinguished (see \cref{sec:MTPD}). Whether a virtual particle is inserted is determined by \cref{eq:virtual_edges} and the particle-triangle distance $d$  (multiple inserts for multiple boundaries).

\begin{figure}[!htb]

\begin{center}
\includegraphics{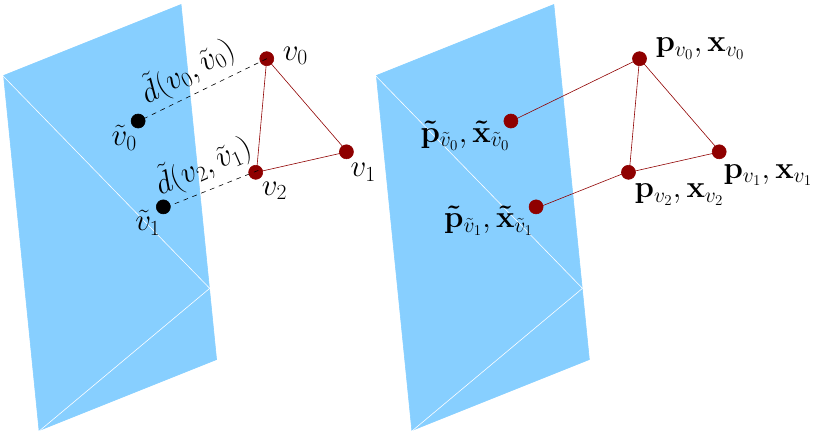}
\end{center}
\caption{Dynamic modification of the graph edges (red lines) and nodes (red points). Left: Calculation of the distances $\tilde d(v_0,\tilde v_0)$, $\tilde d(v_2,\tilde v_1)$ between  real particle at nodes $v_0$, $v_2$ and the triangles corresponding to virtual particle nodes $\tilde v_0$, $\tilde v_1$. Right: Insertion of an additional edge between $\tilde v_0$ and $v_0$ and between $\tilde v_1$ and $v_2$ and representation of the nodes in terms of the corresponding node features $\mathbf{p}_{v_i}$, $\mathbf{x}_{v_i}$ and $\mathbf{\tilde p}_{\tilde v_j}$, $\mathbf{\tilde x}_{\tilde v_j}$ for real and virtual nodes.\label{fig:vars} }	

\end{figure}

\paragraph{Boundary Graph Neural Networks (BGNNs).}
We associate each graph node $v_i$ to a particle with location $\Bx_{v_i}$, 
velocity $\dot{\Bx}_{v_i}$ and acceleration $\ddot{\Bx}_{v_i}$, 
which is similar to~\citet{SanchezGonzalez2020}.
Additionally, we modify and enhance the graph structure to include boundaries (see \cref{fig:vars}).
We dynamically add $\tilde{n}$ virtual nodes $\tilde{v}_j \in \tilde{\cV}$ 
for boundary regions, iff the corresponding boundary region is within a cut-off radius 
to any other particle. 

We augment the set of edges $e_{ij} \in \cE$ by boundary edges $\tilde{e}_{ij} \in \tilde{\cE} $ giving an enhanced edge set $\hat{\cE} = \cE \cup \tilde{\cE}$. 
Analogously to \cref{eq:real_edges1}, the existence of particle-particle edges $e_{ij}$ and  particle-boundary edges $\tilde{e}_{ij}$ is defined via:
{\fontsize{8.0pt}{9.6pt}\selectfont
\begin{align}
    e_{ij} \in \cE \subseteq \hat{\cE} \iff d(v_i, v_j) \leq r_{\text{cut-off}}
    \label{eq:real_edges} \ ,\\
    \tilde{e}_{ij} \in \tilde{\cE} \subseteq
 \hat{\cE} \iff \tilde{d}(v_i, \tilde{v}_j) \leq \tilde{r}_{\text{cut-off}} \ .
    \label{eq:virtual_edges}
\end{align}
}

The cut-off radii $r_{\text{cut-off}}$ and $\tilde{r}_{\text{cut-off}}$ need not necessarily be the same, and,
$d: \cV \times \cV \to \R$, while $\tilde{d}: \cV \times \tilde{\cV} \to \R$, i.e. bidirectional edges are used between real nodes and unidirectional edges are used between real and virtual nodes.
To include information about boundary surfaces into particle-boundary interactions, 
$\tilde{N}$-dimensional node features that encode information about 
the inclination of triangles are concatenated 
with the existing node features $\Bp_{v_i}\in \dR^{N}$.
Additionally, coordinate information is used both for existing nodes
 ($\BX = \{\Bx_{v_0},\ldots,\Bx_{v_{n-1}} \}$) as well as for virtual nodes 
($\tilde{\BX }= \{\tilde{\Bx}_{\tilde{v}_0},\ldots,\tilde{\Bx}_{\tilde{v}_{\tilde{n}-1}} \} $). 
For virtual nodes, the additional coordinates $\tilde{\Bx}_{\tilde{v}_j}$ are chosen such that they minimize the distance between points from boundaries and real particles. The resulting set of node features $\hat{\BP}$ and node coordinates $\hat{\BX}$ are:
{\fontsize{8.0pt}{9.6pt}\selectfont
\begin{align}
\hat{\BP} \ = \ \{\Bp_{v_0},\ldots, \Bp_{v_{n-1}}, \tilde{\Bp}_{\tilde{v}_0},\ldots,\tilde{\Bp}_{\tilde{v}_{\tilde{n}-1}}\}, \  \\  \hat{\BX} \ = \ \{\Bx_{v_0},\ldots,\Bx_{v_{n-1}}, \tilde{\Bx}_{\tilde{v}_0},\ldots,\tilde{\Bx}_{\tilde{v}_{\tilde{n}-1}} \} \  ,
\end{align}
}
\noindent where $\hat{\Bp}_i \in \dR^{N + \tilde{N}}$ and $\hat{\Bx}_i \in \dR^{3}$ denote the elements of $\hat{\BP}$ and $\hat{\BX}$, respectively.
Similarly to above, message passing updates the embeddings of edges ($\hat{\Bm}_{ij}$) and the embeddings of nodes ($\hat{\Bh}_i$) via
{\fontsize{8.0pt}{9.6pt}\selectfont
\begin{align}
\hat{\Bm}'_{ij} \ = \ \hat{\phi}\!\left({\hat{\Bh}}_i, {\hat{\Bh}}_j, 
\hat{\Bm}_{ij}\right), \label{eq:message0} \\
\label{eq:message1}
\hat{\Bh}'_i \ = \ \hat{\psi}\left(\hat{\Bh}_i,\square_{\hat{{e}}_{ij} \in \ \hat{{\cE}}} \  \hat{\Bm}'_{ij} \right)  \ ,
\end{align}
}
\noindent where the aggregation $\square_{\hat{{e}}_{ij} \in \ \hat{{\cE}}}$ at node $v_i$ in \cref{eq:message1} is across all real or virtual nodes that are connected to $v_i$ via an edge $\hat{e}_{ij}$.
Similar to \citet{Gilmer2017neural} and \citet{Satorras2021egnn}, we make use of pairwise distances ($\NRM{\hat{\Bx}_i-\hat{\Bx}_j}^2$ and $\hat{\Bx}_i -\hat{\Bx}_j$ and deterministic functions thereof). These are for BGNNs between real and between real and virtual particles and we pass this information to the graph network as edge attributes $\hat{\Ba}_{ij}$, for which an initial edge embedding $\hat{\Bm}_{ij}$ is determined via an edge embedding layer.
The final node embeddings are used for the predictions via the read-out layers. For aggregation $\square$, we use the mean.

\paragraph{Dynamical graph model.}
At each time point a graph of the current scene is built up, containing the minimum distances between particles and walls as well as distances between particles within certain neighborhoods. The definition of the graph and computations on it make up our effective theory. Especially, every particle ``sees'' at most one virtual particle representing the boundary surface area, namely that virtual particle which has the shortest distance. 
\Cref{tab:particle_comparison} shows average numbers of nodes $\big|\cV\big|$, as well as average numbers of boundary edges  $\big|\tilde{\cE}\big|$ and the relative increase in edges (ratio of the number of added wall edges to the total number of particle edges). 
The scalability of BGNNs would suffer if more than one particle per particle-boundary interaction surface was considered. Our approach is summarized in \cref{algBGNN}.

\begin{algorithm}[!htb]
\caption{\label{algBGNN} BGNN: Dynamic Graph Message Passing}
\begin{algorithmic}[1]

\STATE $\text{fr}_{s,t} \gets $ random frame from trajectory $s$ at time $t$
\STATE Assign node $v_{i}$ to each real particle at $\Bx_{v_i}$ with features $\Bp_{v_i}$ and calculate pairwise distances $d_{ij}=d(v_i, v_j)$ between nodes $v_i$ and $v_j$. Assign edges $e_{ij}$ with edge features to those fulfilling \cref{eq:real_edges}.
\STATE Calculate $\tilde{d}_{ij}=q\left(u_{0;(v_i,\triangle_j)}',u_{1;(v_i,\triangle_j)}'\right)$ for all particles ($v_i$) and triangles ($\triangle_j$) according to \cref{eq:triPointMin}. In case \cref{eq:virtual_edges} is fulfilled, insert (a) a virtual node $\tilde{v}_j$ at $\tilde{\Bx}_{\tilde{v}_j}=\Bt\left(u_{0;(v_i,\triangle_j)}',u_{1;(v_i,\triangle_j)}'\right)$ with associated triangle-specific features according to \cref{sec:orInd}, and (b) an edge between the real and the inserted virtual node together with associated edge features.
\STATE Fill up empty triangle-specific real-node features $\Bp_{v_i}$ and particle-specific virtual-node features $\tilde{\Bp}_{\tilde{v}_{\tilde j}}$ with null values and add indicator features for node types (real/virtual) to obtain $\hat{\Bp}_i$ and $\hat{\Bp}_j$.
\STATE Apply BGNN message passing according to \cref{eq:message0,,eq:message1}.
\STATE Update positions via semi-implicit Euler method. 

\end{algorithmic}
\end{algorithm}

\begin{table}[!htb]
\centering

\footnotesize
\begin{tabular}{c  r@{\hspace{0.1cm}}l r@{\hspace{0.1cm}}l c}

\toprule
Experiment  & \multicolumn{2}{c}{ $\displaystyle \big|\cV\big|$} &  \multicolumn{2}{c}{ $\displaystyle \big|\tilde{\cE}\big|$} & $\%$ increase \\
\midrule
Hopper & 1113 &  $\pm$ 738 & 5475 &  $\pm$ 3547 & 72.2\\
Drum & 3283 &  $\pm$ 282 & 1678 &  $\pm$ 188 & 54.8\\
\bottomrule
\end{tabular}
\caption{ Growth of the number of edges due to boundaries in the graph. The table shows statistics across the training trajectories of non-cohesive particles in a standard setting in hopper and drum experiments. For each trajectory the frame with maximum relative increase in the number of edges due to virtual particles $\displaystyle  \protect \big|\tilde{\cE}\protect \big|\protect \big/ \protect \big|\cE \protect \big|$ has been selected as a representative frame. This is done since we are interested in the maximum effect additional virtual particles have on the memory requirements. Number of particles $\protect \big|\cV \protect \big|$, number of additional virtual edges $\protect \big|\tilde{\cE} \protect \big|$, and $\%$ increase are listed.\\
}\label{tab:particle_comparison}
\end{table}

\paragraph{Boundary normal directions.}
\label{bfeat}
Typical granular flow simulations comprise substantially more particle-particle interactions than particle-boundary interactions, which may impede the learning of particle-boundary interactions. In \citet{kipf2018neural} the problem of qualitatively different interactions is addressed by introducing a dedicated message generating network for each interaction type. We avoid such extensions of our model by means of the following two approaches.
First, we introduce additional node features, such that the neural network is able to distinguish the different types of nodes.
Second, we adapt the weight initialization of the node feature embedding $\hat{\psi}$, such that the embedding network can be trained with larger values for the additional features.
Consequently, the network can learn different dynamics for particle-particle and particle-boundary interactions. The additional node features are: (i) type feature, i.e.,  a binary indicator of whether a node represents a particle that is real or virtual, and, in the latter case, (ii) the components of the normal vector  (see \cref{sec:orInd} for more information on an orientation-independent representation of the normal vectors) of the triangular surface areas (null vectors for real particles).

\section{Related Work}

There is a rich body of literature on applications of Deep Learning in the context of physics simulations. 
Most notably related to BGNNs are the works of \citet{SanchezGonzalez2020}, \citet{ummenhofer2019}, and, \citet{li2018learning}, all of which propose methods of learning particle simulations without enforcing constraints. These approaches can be contrasted to works like \citet{ladicky2015data} or \citet{schenck2018spnets} that utilize strong inductive biases. \citet{ladicky2015data} construct features for Random Forest Regression that are influenced by Smooth Particle Hydrodynamics \citep{Gingold1977, Lucy77}. 
\citet{schenck2018spnets} construct a differentiable fluid dynamics network that is closely related to the Position Based Fluids method \citep{macklin2013position}. Importantly, both methods are built on the assumption that the governing equations of the system are known, which is, as mentioned in \cref{sec:bck}, not necessarily the case for granular flow dynamics.

Integrating finite element methods and therefore triangularized boundaries into deep learning architectures has started to gain interest~\citep{longo2022rham}.
Complex mesh-based wall geometries have been employed 
to compute updates for nodes of the mesh itself \citep{Pfaff2020}.
In contrast to \citet{Pfaff2020} in our scenario, the mesh is static, i.e. the descriptive representation of machine parts. 
We share the opinion of \citet{SanchezGonzalez2020} that the network architecture with continuous convolutions as suggested by \citet{ummenhofer2019} can be interpreted as GNNs. 
In doing so, a difference to \citet{SanchezGonzalez2020} and our work is that \citet{ummenhofer2019} use static particles as special nodes in the first message passing step only.
Consequently, the framework of \citet{SanchezGonzalez2020}, which is based on \citet{Battaglia2018}, appears to be the most general to us, performing well even without explicit hierarchical clustering as suggested in DPI-Net \citep{li2018learning}. 
Experiments of \citet{SanchezGonzalez2020} further suggest that their simulation of sand particles are superior to the implementation of \citet{ummenhofer2019}.
However, \citet{SanchezGonzalez2020} only consider simple cuboid boundaries for their 3D simulations, leaving more realistic complex geometries as an open and yet untouched challenge.
Furthermore, they use sampled, static particles to represent boundaries for 2D simulations, which in general does not scale well for 3D simulations due to the quadratic increase of boundary particles (square areas instead of lines).

\section{Experiments}
\label{sec:experiments}
\begin{figure}[!ht]
\begin{center}
\includegraphics{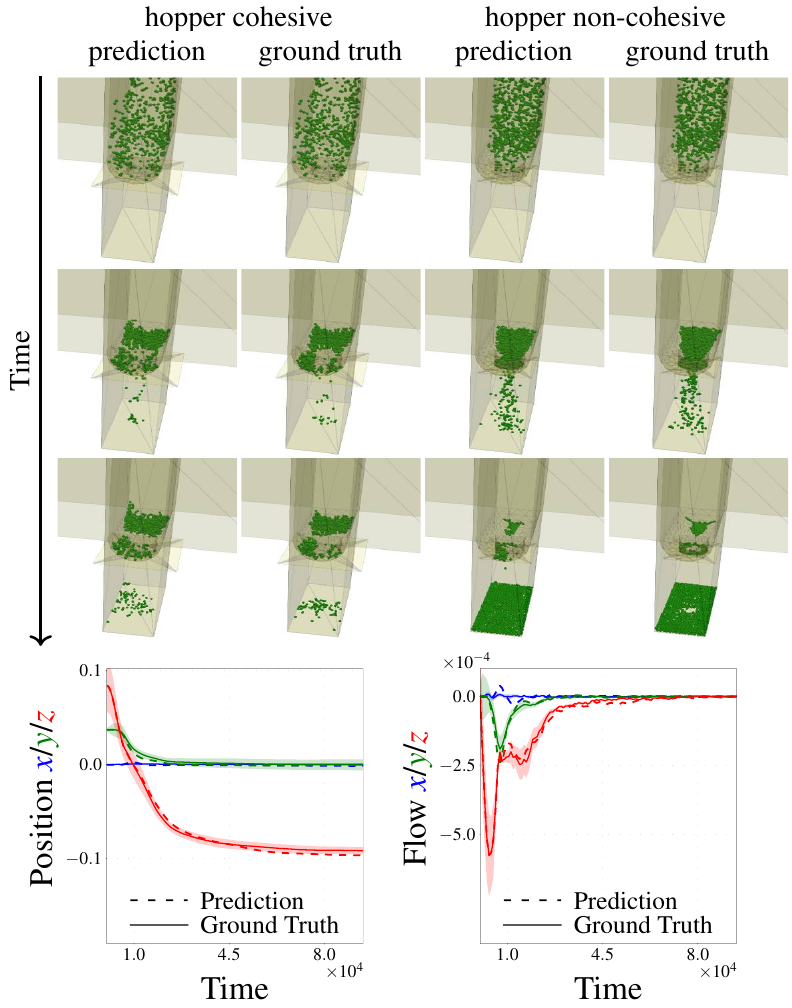}
\end{center}
\caption{ Hopper dynamics. Top: Distributions for cohesive and non-cohesive particles. Simulation data and BGNN predictions are compared. Particles are indicated by green spheres, triangular wall areas are yellow, the edges of these triangles are indicated by grey lines. In contrast to liquid-like non-cohesive particles, cohesive particles lead to congestion of the hopper. 
Bottom: Position (left) and flow profile (right) for non-cohesive particles. Corresponding plots for cohesive particles can be found in \cref{sec:app_experiments}. Simulation data (solid lines) and BGNN predictions (dashed lines) are compared. Simulation uncertainties are due to a change of the particle numbers ($\pm25\%$) and to different initial conditions. We provide simulation predictions for a hopper with more timesteps in animations at \url{https://ml-jku.github.io/bgnn/}.}
\label{fig:hopper_results}
\end{figure}

\begin{figure}[!ht]
\begin{center}
\includegraphics{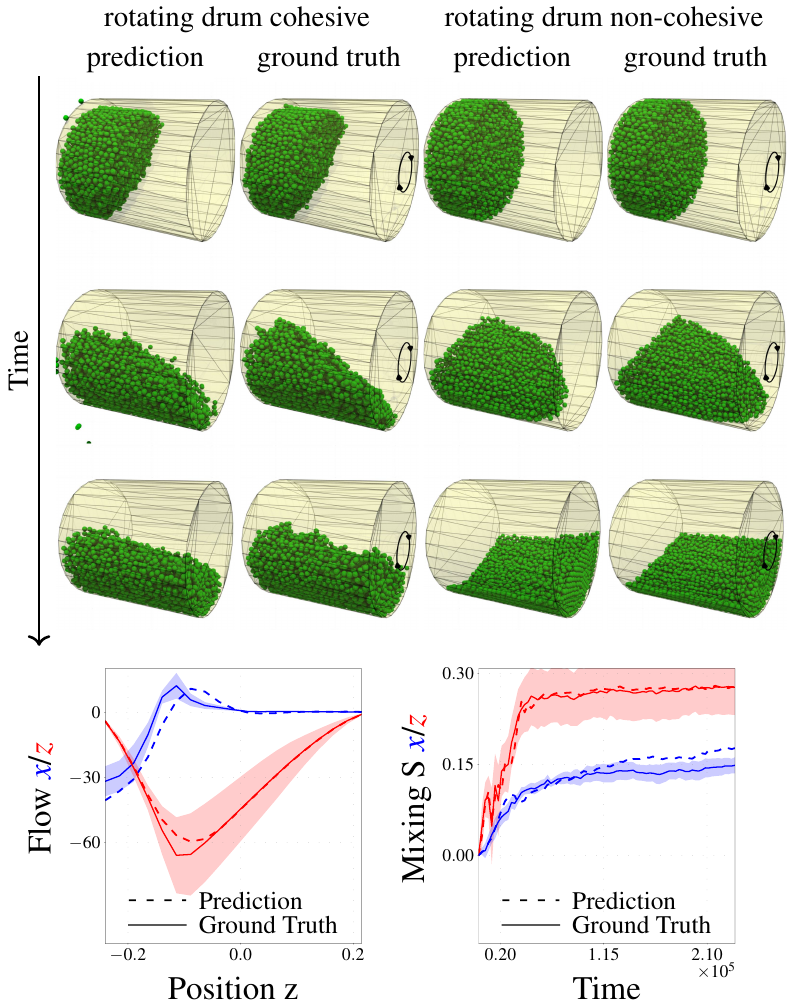}
\end{center}
\caption{ Rotating drum dynamics. Top: Particle distributions for cohesive and non-cohesive particles. Simulation data and BGNN predictions are compared. Particles are indicated by green spheres, triangular wall areas are yellow, the edges of these triangles are indicated by grey lines.
The circular arrow indicates the rotation direction of the drum.
In contrast to liquid-like non-cohesive particles, 
cohesive particles stick together much stronger.
Bottom: Flow profile (left) and entropy plot (right) for
non-cohesive particles. The entropy is shown for particle class assignment according to the x (blue) and z (red) position.
Corresponding plots for cohesive particles can be found in \cref{sec:app_experiments}.
Simulation data (solid lines) and BGNN predictions (dashed lines) are compared. Simulation uncertainties are due to a change of the particle numbers ($\pm25\%$) and to different initial conditions. We provide simulation predictions for a rotating drum with more timesteps in animations at \url{https://ml-jku.github.io/bgnn/}.
}	
\label{fig:rotating_drum_results}
\end{figure}

We test the effectiveness of BGNNs 
on complex 3D granular flow simulations.
The development, design, and construction of many mechanical devices is based on granular flow simulations.
These devices can have very different geometries and must 
be designed for a wide range of materials with highly varying properties. 
For example, cohesion properties can range from dry, wet, to oily. 
In the simulations, 
we consider very common device geometries and different
cohesion properties, as well as static and moving geometries, to cover a wide range of situations 
with our available computational resources.
The two common geometries are hoppers and rotating drums
(see \cref{fig:example_triangularization,,fig:hopper_results,,fig:rotating_drum_results}). 
The two different cohesion properties are non-cohesive
describing liquid-like, oily materials and cohesive describing
dry, sand-like materials.
We compare the BGNN predictions to the simulations in two aspects: 
speed and accuracy.

\paragraph{Simulation Details. }
For all experiments, gravitation acts along the $z$-direction. 
The upper part of the hopper is delimited along the $y$-axis by two planes, 
which are parallel to the $x$-$z$ plane (see \cref{fig:hopper_results}). 
The $x$-axis is delimited by two planes, that are inclined 
at certain angles $\alpha,\, 180^{\circ}-\alpha$ to the $x$-$y$ plane and 
at corresponding angles $\alpha-90^{\circ}, 90^{\circ}-\alpha$ 
to the $y$-$z$ plane. 
The hopper has an initially closed hole at the bottom, 
which has an adjustable radius.
The rotation axis of the drum is the $y$-axis 
(see \cref{fig:rotating_drum_results}).
The initial filling of the hopper and drum is done by randomly inserting particles into a predefined region, see \cref{sec:app_experiments}.
We use around $1000$ and around $3000$ particles for hopper and rotating drum simulations, respectively. 
In order to have trajectories with non-cohesive and cohesive particles, we use the simplified JKR model \citep{roessler2019parameter} with a cohesion energy density of $0$ J/$m^3$ and $10^5$ J/$m^3$ for non-cohesive and  
cohesive particles.
The training data consists of 30 simulation trajectories,
where each trajectory consists of 100.000 (250.000) simulation timesteps for hopper (rotating drum). 
For BGNN training every 40 (100)-th timestep is used.
Trajectories have different angles $\alpha$ and different hole radii (hopper) and different initial particle placement (drum). Moreover, the number of particles is varied by $\pm 25 \%$. 

\paragraph{Implementation Details. }
We use 3 to 10 message passing layers, with 128 and 512 nodes for intermediate node and edge representation. The cut-off radii strongly depend on the particle size. We use cut-off radii of 0.02 and 0.008 for rotating drum and hopper, respectively. Cut-off radii have been treated as hyperparameters of our model.
More details can be found in \cref{sec:app_experiments}.

\paragraph{Assessment Of Physical Quantities. }
Granular flow simulations should correctly describe systems on macroscopic scales in terms of
\textit{particle-averaged positions} $\bar{\mathbf{x}}(t)$ 
and \textit{particle flows} $\bar{\mathbf{v}}(t)$ 
for $n$ particles as a function of time: 
$\bar{\mathbf{x}}(t) = \frac{1}{n}\sum_i \mathbf{x}_i(t)$ and
$\bar{\mathbf{v}}(t) = \frac{1}{n} \sum_i \mathbf{v}_i(t)$.
Hoppers are devices that aim at adjusting the flow of particles 
along the direction of gravity, 
which coincides with the $z$-axis in our experiments.
Rotating drums are commonly utilized as mixing devices for various applications in e.g. industry, research, and agriculture. 
They are essentially rotating cylinders that are partially filled 
with a granular material. 
The mixing property of these devices is a result of 
numerous particle interactions under time-varying boundary conditions.
For rotating drum experiments, we quantify the extend of particle mixing via the \textit{mixing entropy}~\citep{Fang1975appl}. 
If the z-coordinate of a particle's initial position $\mathbf{x}_i(0)$
is above (below) the median z-coordinate of all particles in the initial state, we assign it to class $c=+1\ (-1)$. Based on this assignment local entropies $s(\Bg_{klm}, t)$ at grid cells $\Bg_{klm}$ are calculated, where the indices $klm$ identify an individual grid cell. The local entropies $s(\Bg_{klm}, t)$ are computed from particle counts $n_{c}(\Bg_{klm}, t),$ of the respective classes $c=\pm1$. The total number of particles in a grid cell is obtained by $n(\Bg_{klm}, t)=n_{+1}(\Bg_{klm}, t)+n_{-1}(\Bg_{klm}, t)$. Calculating the particle-number weighted average of the local mixing entropies yields the mixing entropy  $\rS(t)$ of the entire system:
{\fontsize{8.0pt}{9.6pt}\selectfont
\begin{align} \label{eq:eq_entropy}
\begin{split}
\rS(t)  &=  \frac{- \ \sum_{klm} \sum\limits_{c=\pm1} \ n(\Bg_{klm}, t) \ \big(f_{c}(\Bg_{klm}, t) \log f_{c}(\Bg_{klm}, t)\big)}{\sum\limits_{klm}  n(\Bg_{klm}, t)}  , 
\end{split} \nonumber
\end{align}
} \noindent where $f_{c}(\Bg_{klm}, t)$ denotes the relative fraction of class $c$ particles in cell $\Bg_{klm}$ at time $t$.

\paragraph{Results. }
In \cref{fig:hopper_results} and \cref{fig:rotating_drum_results}
results for the hopper and the rotating drum simulations are presented. 
The upper parts visualize granular flow snapshots 
at different time steps, both for cohesive and non-cohesive materials. 
The lower parts of the figures include average position and 
particle flow plots for hopper, 
as well as particle flow and mixing entropy plots for rotating drum simulations. 
The simulation uncertainties arise due to the different distributions 
of the initial filling and due to a $\pm 25\%$ variation 
in the number of particles across simulations. 
The difference between cohesive and non-cohesive particles is evident. BGNNs have learned to model granular flow simulations 
over thousands of time steps. 
Most notably, hardly any particle leaves the geometric boundaries. 
This is achieved without using handcrafted conditions or restrictions 
on the positions of the particles. 
Furthermore, BGNNs have learned to model particle-boundary interactions and 
in doing so correctly represent the dynamics within the system. 
The predicted quantities are within uncertainties of the simulations. 

\begin{figure}[!ht]
\begin{center}
\includegraphics{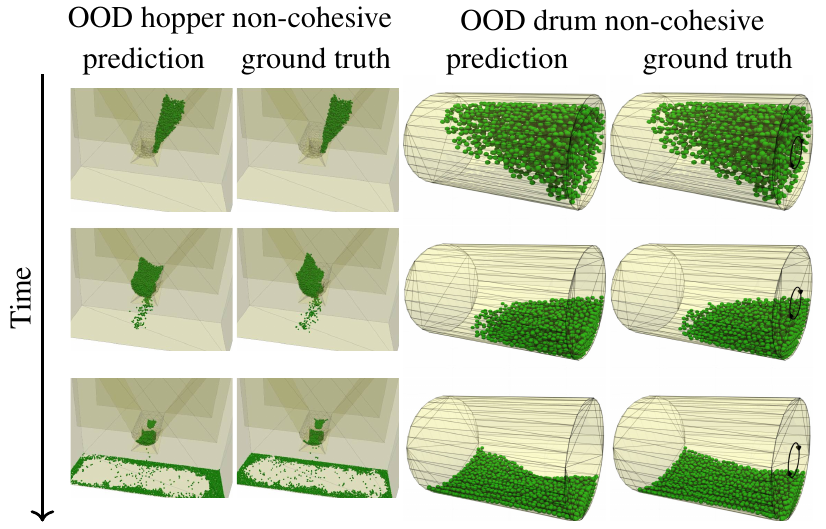}
\end{center}
\caption{OOD generalization behavior for the hopper (left) and the rotating drum (right). In contrast to the training and validation data the outlet size of the hopper was decreased, the inclination angles of the hopper side walls are enlarged, and, the length of the rotating drum is increased.}	
\label{fig:ood}

\end{figure}

\paragraph{Out-of-distribution generalization.}
Therefore, we consider the BGNN predictions as sufficiently precise to substitute the simulations. \Cref{fig:ood} shows out-of-distribution (OOD) scenarios, where the devices are changed with respect to the training data. The hole size of the hopper is

\begin{figure}[!ht]
\begin{center}
\includegraphics{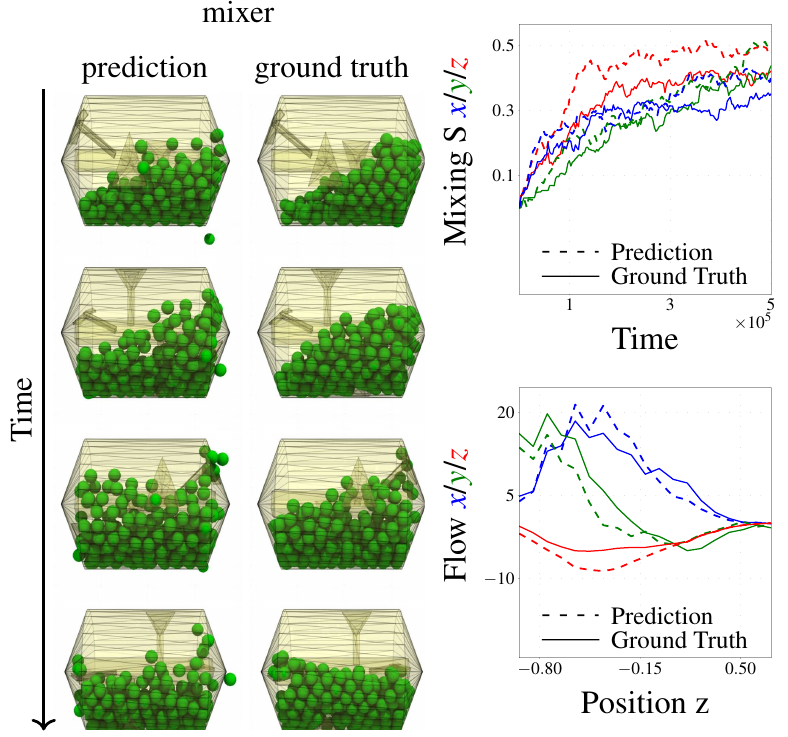}
\end{center}
\caption{Mixer dynamics. A mixer can be seen as a moving geometry where additional difficulty is imposed due to rotating blades inside. BGNNs are also well suited to model such scenarios of increased difficulty. 
}	
\label{fig:mixer_results}
\end{figure}

decreased in mean by $\sim$ 50\%, while side wall inclination angles have been increased by $\sim$ 15°. For the drum the 
length of the corresponding cylinder was increased in mean by $\sim$ 50\%. Our experiments show that our model generalizes well across variations in the geometry. This finding demonstrates that trained BGNNs could be used for designing and studying different geometries without retraining the model. 

\paragraph{Moving geometries.}
As an additional challenge we consider moving geometries, as shown in \cref{fig:mixer_results}, where additional difficulty is imposed due to a rotating blade inside a particle mixer. Consequently, not only the geometry but also the blade itself are triangularized and particle-surface interactions are extended by particle-blade interactions. Experiments show that our BGNN approach is well suited to model such scenarios of increased difficulty.

\paragraph{Runtime}
\Cref{tab:runtime_comparison} gives a run-time comparison 
of the LIGGGHTS simulation versus a forward pass of BGNNs,
which only predict every 100 time step. 
The highly optimized CPU algorithm (LIGGGHTS) and 
a non-optimized GPU compatible algorithm (BGNNs) are compared
via their wall-clock times since the hardware settings are 
quite different.
Nevertheless, \cref{tab:runtime_comparison} shows that 
the wall-clock time of BGNNs is shorter than the wall-clock time
of the simulation. 
The usage of more particles, would further increase the lead of BGNNs
over the simulation in terms of wall-clock time.
For the time comparison, we use a typical simulation trajectory 
from our datasets with 3,408 particles, which needs approximately 2 GB GPU memory for one forward pass.
An essential reason for speedup in our (simple) setups are GPU parallelization capabilities.
There is potentially even more space for improvement of the BGNN predictions over simulations due to the so called \textit{Young's modulus}.
For simulations, it is often assumed that energy
is purely transmitted through Rayleigh waves. 
Thus the time step of DEM simulations is targeted 
to be a fraction of the propagation time through a single, solid particle. As such the propagation time depends on material parameters, most notably the Young's modulus. 
However, for several materials the Young's moduli 
that reflect the true material properties, 
would lead to extremely small propagation times, which in turn means much more simulation steps. 
Consequently, much smaller Young's moduli are considered as an approximation, which is valid for gravity driven flows \citep{coetzee2017calibration}.
However, for many cases, e.g. the penetration of a particle bed by an object, this approximation breaks down~\citep{lommen2014speedup}.
BGNNs have the potential to be trained on very small time steps 
reflecting the true Young's moduli and consequently 
generalize over much more than ``just'' 40 or 100 time steps.

\begin{table}[!h]
\centering
\footnotesize
\begin{tabular}{l c c c c}
\toprule
method & \makecell{time steps} & \makecell{real world \\ time} &  \makecell{ wall-clock \\ time [$s$]} \\
\midrule
\vspace{0.1cm}
LIGGGHTS & $250.000$ & $12.5s$  & $356$ \\
BGNNs & $2.500$ & $12.5s$ &  $158$\\
\bottomrule
\end{tabular}
\caption{Runtime comparison for one granular flow process (rotating drum) consisting of 250.000 simulation timesteps, which are 2500 BGNN predictions. For both, simulation and BGNN trajectories, this corresponds to a real world time of $12.5s$. LIGGGGHTS simulation is run on a \shortstack{CPU AMD EPYC 7H12}, BGNN forward pass is run on a \shortstack{GPU NVIDIA A100}.\\}
\label{tab:runtime_comparison}
\end{table}

\section{Conclusion and Future Directions}

We have introduced an effective theory to model complex particle-boundary interactions, resulting in Boundary Graph Neural Networks (BGNNs).
BGNNs dynamically modify graph structures via modifying edges, augmenting node features, and dynamically inserting virtual nodes.
BGNNs achieve an accurate neural network modeling of simulated physical processes within complex geometries.
We have tested BGNNs on complex 3D granular flow processes of hoppers, rotating drums, and mixers, where BGNNs are able to accurately reproduce these flows 
within simulation uncertainties over hundreds of thousands of timesteps. Most notably particles stay within the geometric objects without using handcrafted conditions or restrictions.
A possible extension of our work is towards a wide range of different materials, e.g. materials with high Young's moduli as described in \cref{sec:experiments}. 
Another interesting extension is to introduce a velocity dependent cut-off radius, and in doing so considering also those particle-boundary interactions which are about to happen within the next timesteps although the spatial distances are still large. 
Finally, leveraging the symmetries and geometries of granular flow problems~\citep{brandstetter2021geometric, brandstetter2022clifford} is appealing. 

\section*{Availability}

Code and further information can be found at
\url{https://ml-jku.github.io/bgnn/}\\

\section*{Acknowledgments}

This research was supported by FFG grant 871302 (DL for GranularFlow). 

The ELLIS Unit Linz, the LIT AI Lab, the Institute for Machine Learning, are supported by the Federal State Upper Austria. IARAI is supported by Here Technologies. We thank the projects AI-MOTION (LIT-2018-6-YOU-212), DeepFlood (LIT-2019-8-YOU-213), Medical Cognitive Computing Center (MC3), INCONTROL-RL (FFG-881064), PRIMAL (FFG-873979), S3AI (FFG-872172), EPILEPSIA (FFG-892171), AIRI FG 9-N (FWF-36284, FWF-36235), ELISE (H2020-ICT-2019-3 ID: 951847), Stars4Waters (HORIZON-CL6-2021-CLIMATE-01-01). We thank Audi.JKU Deep Learning Center, TGW LOGISTICS GROUP GMBH, Silicon Austria Labs (SAL), FILL Gesellschaft mbH, Anyline GmbH, Google, ZF Friedrichshafen AG, Robert Bosch GmbH, UCB Biopharma SRL, Merck Healthcare KGaA, Verbund AG, GLS (Univ. Waterloo), Software Competence Center Hagenberg GmbH, T\"{U}V Austria, Frauscher Sensonic and the NVIDIA Corporation.

We thank Angela Bitto-Nemling, Markus Holzleitner, and, G\"unter Klambauer for helpful discussions and comments on this work.

\bibliographystyle {aaai23}
\bibliography{refsMain}

\appendix

\renewcommand{\thesection}{TApp. \Alph{section}}

\onecolumn

{\centering {\LARGE\bf Technical Appendix: \\ Boundary Graph Neural Networks for 3D Simulations \par }}
\vskip 1in

\numberwithin{equation}{section}
\numberwithin{figure}{section}
\numberwithin{table}{section}

\imagelabelset{
coarse grid color = red,
fine grid color = gray,
image label font = \sffamily\small,
image label distance = 2mm,
image label back = black,
image label text = white,
coordinate label font = \normalfont,
coordinate label distance = 2mm,
coordinate label back = white,
coordinate label text = black,
annotation font = \normalfont\small,
arrow distance = 1.5mm,
border thickness = 0.6pt,
arrow thickness = 0.4pt,
tip size = 1.2mm,
outer dist = 0.5cm,
}

\FPeval{\prop}{0.45}
\FPeval{\coordOn}{\prop*0.25}
\FPeval{\coordTw}{\prop*0.75}
\FPeval{\coordOnTw}{\prop*0.5}
\FPeval{\coordTh}{\prop+(1-\prop)*0.25}
\FPeval{\coordFo}{\prop+(1-\prop)*0.75}
\FPeval{\coordThFo}{\prop+(1-\prop)*0.5}

\section{Discrete Element Method (DEM) simulator LIGGGHTS}
\label{sec:DEM}

\begin{figure}[h]
\begin{center}
\includegraphics[width=0.21\textwidth, trim=150 30 350 10, clip=True]{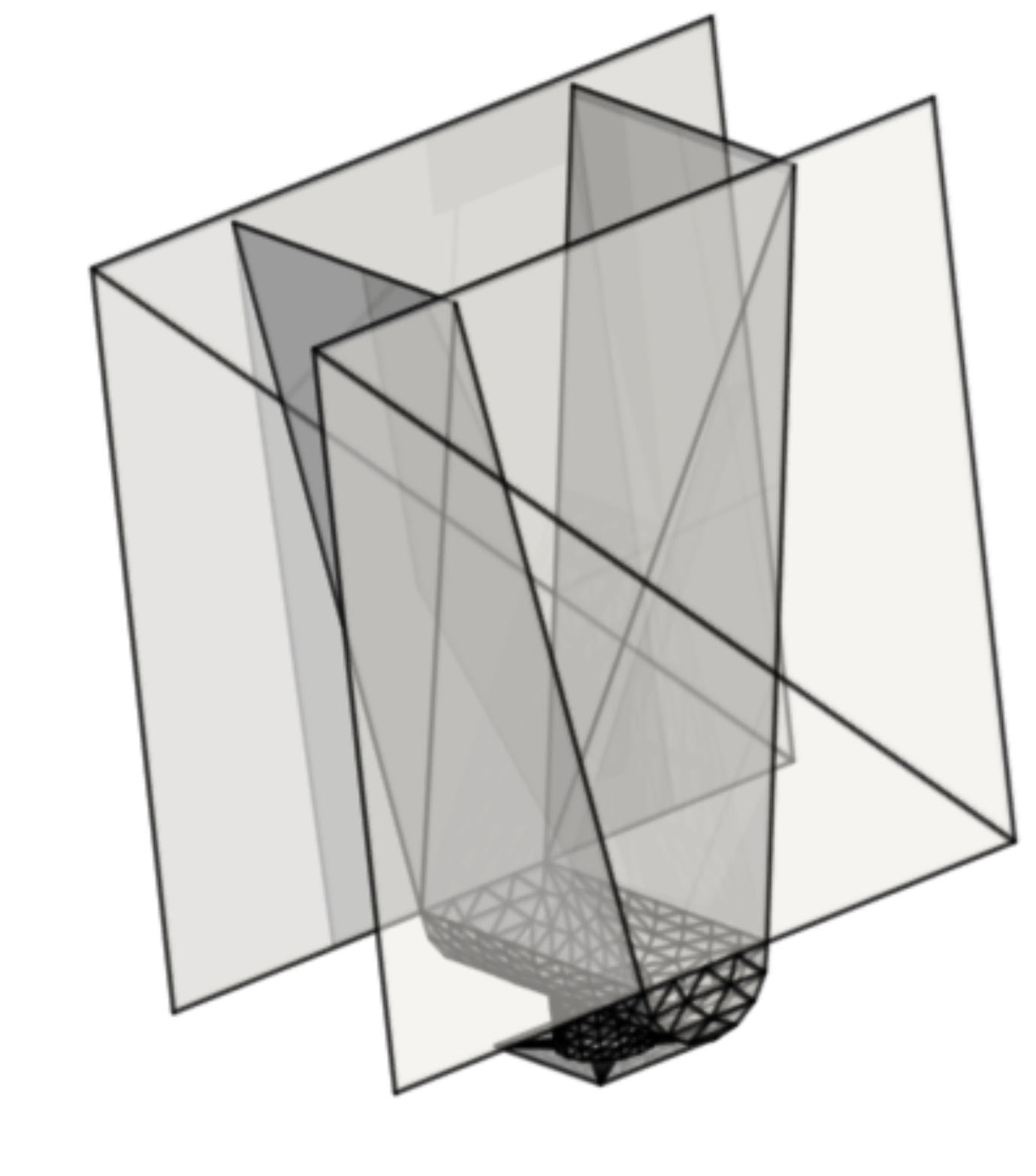} \qquad
\includegraphics[width=0.21\textwidth, trim=250 20 250 50, clip=True]{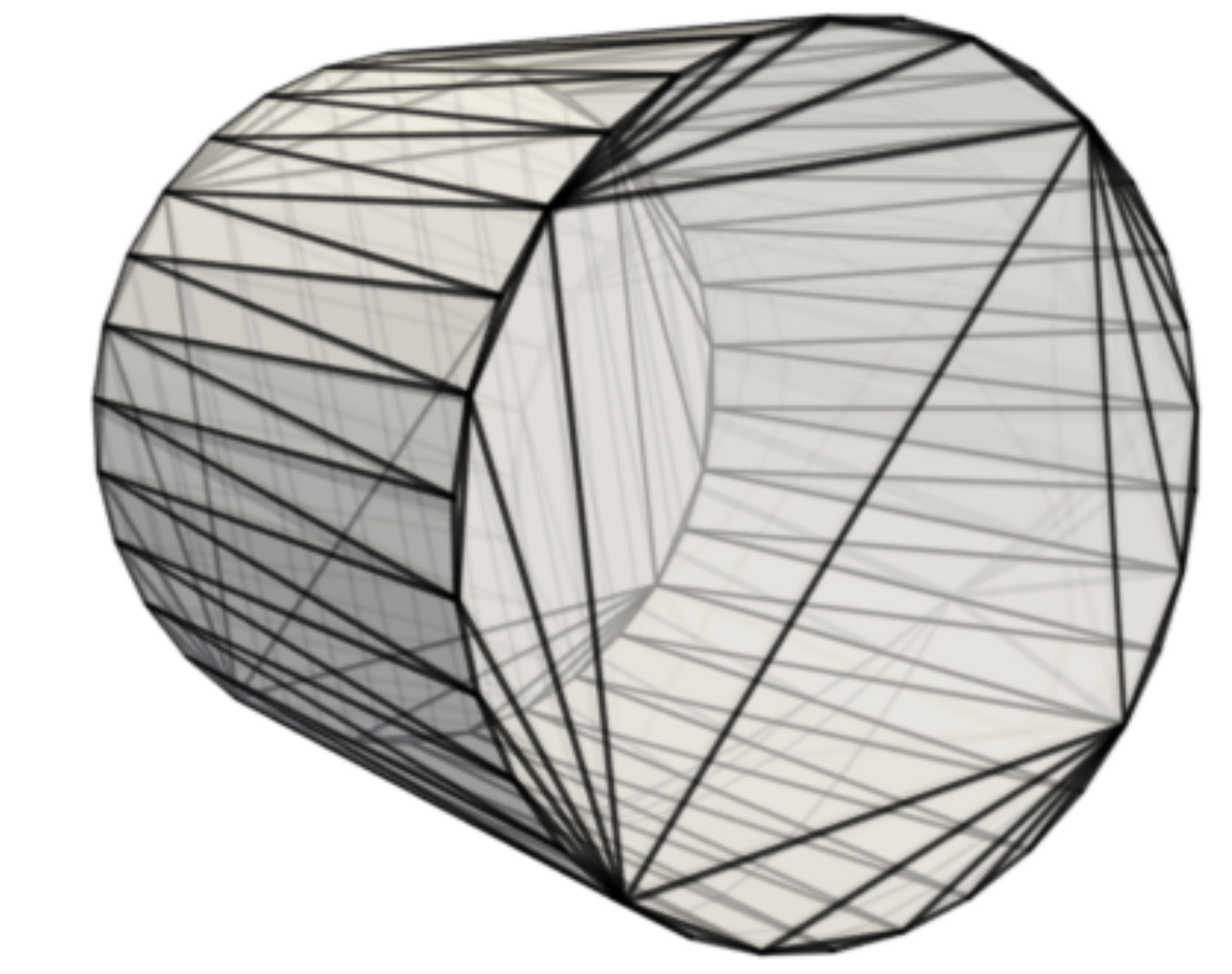}
\end{center}
\caption{Triangularized boundary surfaces of the hopper (left) and the rotating drum (right). The accurate description of rather simple, curved geometries requires a relatively large number of triangles of different shapes and sizes.
\label{fig:example_triangularization}} 
\end{figure}

The open source DEM software LIGGGHTS \citep{Kloss2012} is based on the Molecular Dynamics code LAMMPS \citep{plimpton1995fast} developed by Sandia National Labs.
Due to the similarity of the underlying algorithms for neighbor list construction, output and parallelism this provided a stable basis for the contact models required for DEM.
LIGGGHTS added support for triangular mesh walls (as e.g. such ones visualized in \cref{fig:example_triangularization}), particle insertion and new particle shapes (multispheres and superquadrics).
Several of those changes resulted in upstream contributions in LAMMPS.

Over the years LIGGGHTS has become a widely used software in both academia and industry that supports both cutting edge research and industrial applications.
Support for several physical phenomena as, e.g. liquid transfer on particles, was instrumental in its success.
However, it also highlighted requirements for additional research.
In industrial applications there often is the need to study physical phenomena which occur on different time scales, e.g.\ particle collisions ($\mathcal{O}(10^{-5}s)$) vs. moisture content in particles ($\mathcal{O}(1s)$, \citep{Mellmann2011}), which can lead to weeks of simulation time.
While advances have been made to overcome such issues (e.g.\ \citet{Kloss2017}), they remain limited in their application, due to the fact that they rely on prior simulation of the exact setup and cannot be used for interpolation of quantities directly related to the flow behavior.

In 2019 LIGGGHTS was again forked and forms the basis of the commercial DEM software Aspherix\textregistered\ which expands the capabilities of LIGGGHTS with polyhedral particles, a significantly simplified input language and graphical user interface.

\section{Minimum Triangle - Point Distance}
\label{sec:MTPD}

\begin{figure}[!ht]
\begin{center}
\includegraphics[trim=0 10 0 0, scale=0.45]{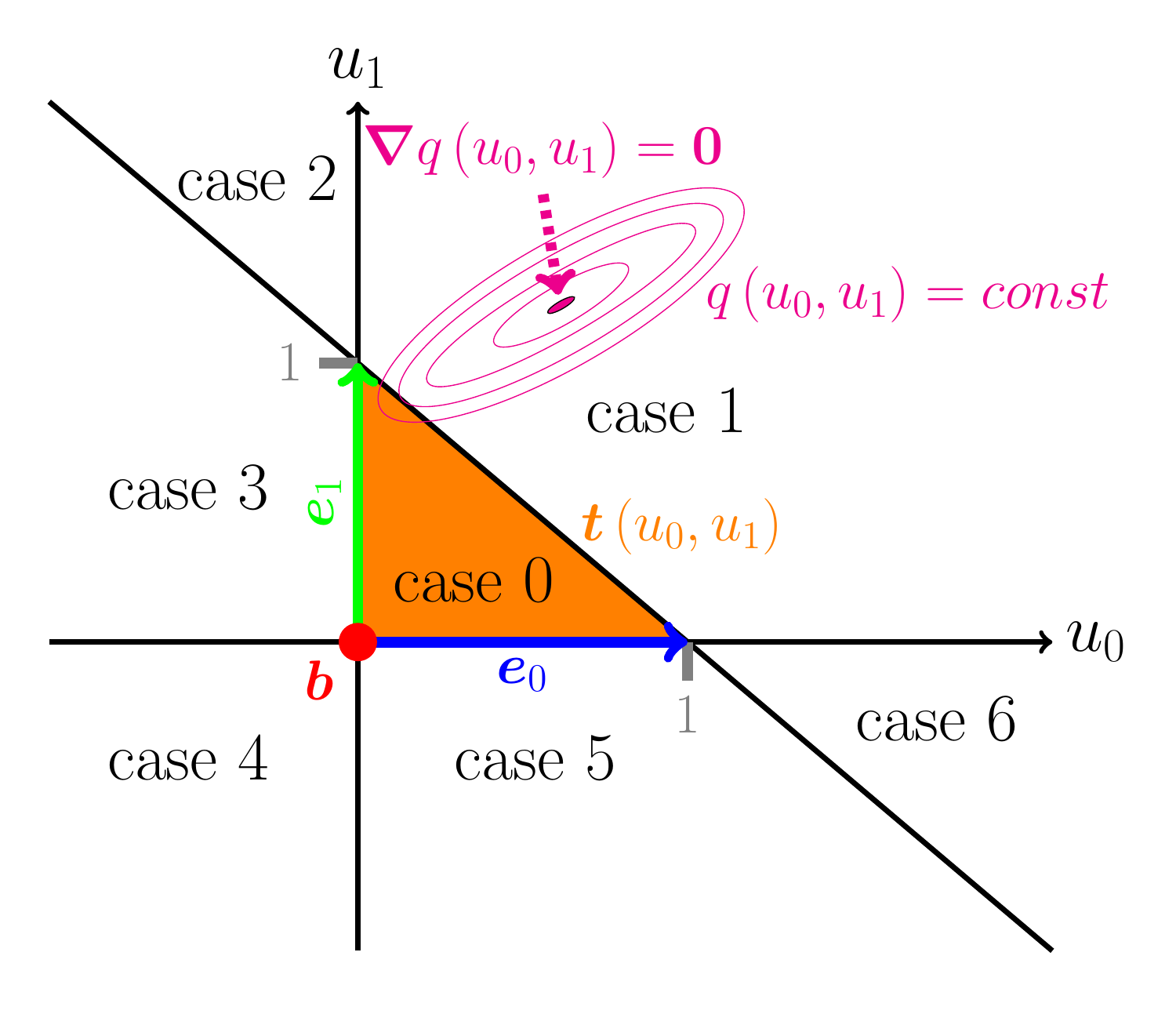}
\end{center}
\caption{Visualization of point - triangle distance calculations in 3D. The triangle is represented by a parameterized function $\Bt\left(u_0,u_1\right)=\Bb+u_0\;\Be_0+u_1\;\Be_1$ with $u_0\geq0$, $u_1\geq0$, $u_0+u_1\leq1$ (indicated by the orange area). Level sets of $q\left(u_0,u_1\right)$ are indicated by ellipses and describe the squared Euclidean distance of a triangle point $\Bt\left(u_0,u_1\right)$ to the point $\Bp$, for which we compute the minimum distance. \label{fig:myTriangle} } 
\end{figure}

The main part of the manuscript states the
optimization problem, which needs to be solved.
According to \citet{Eberly1999}
seven cases have to be to distinguished: 
one (c0) in which $\Bt(u_0',u_1')$ is located within the (closed) triangle, 
three (c1, c3, c5) in which $\Bt(u_0',u_1')$ is located on one edge of the triangle (including the edge corner points as special cases),
and three (c2, c4, c6) in which $\Bt(u_0',u_1')$ is located on one of two edges (including the triangle corner points as special cases).
We visualize these cases in \cref{fig:myTriangle}.

\section{Normal Vector Representations}
\label{sec:orInd}

\begin{figure}[ht]
\begin{center}
\includegraphics[trim=130 0 0 0, scale=0.27]{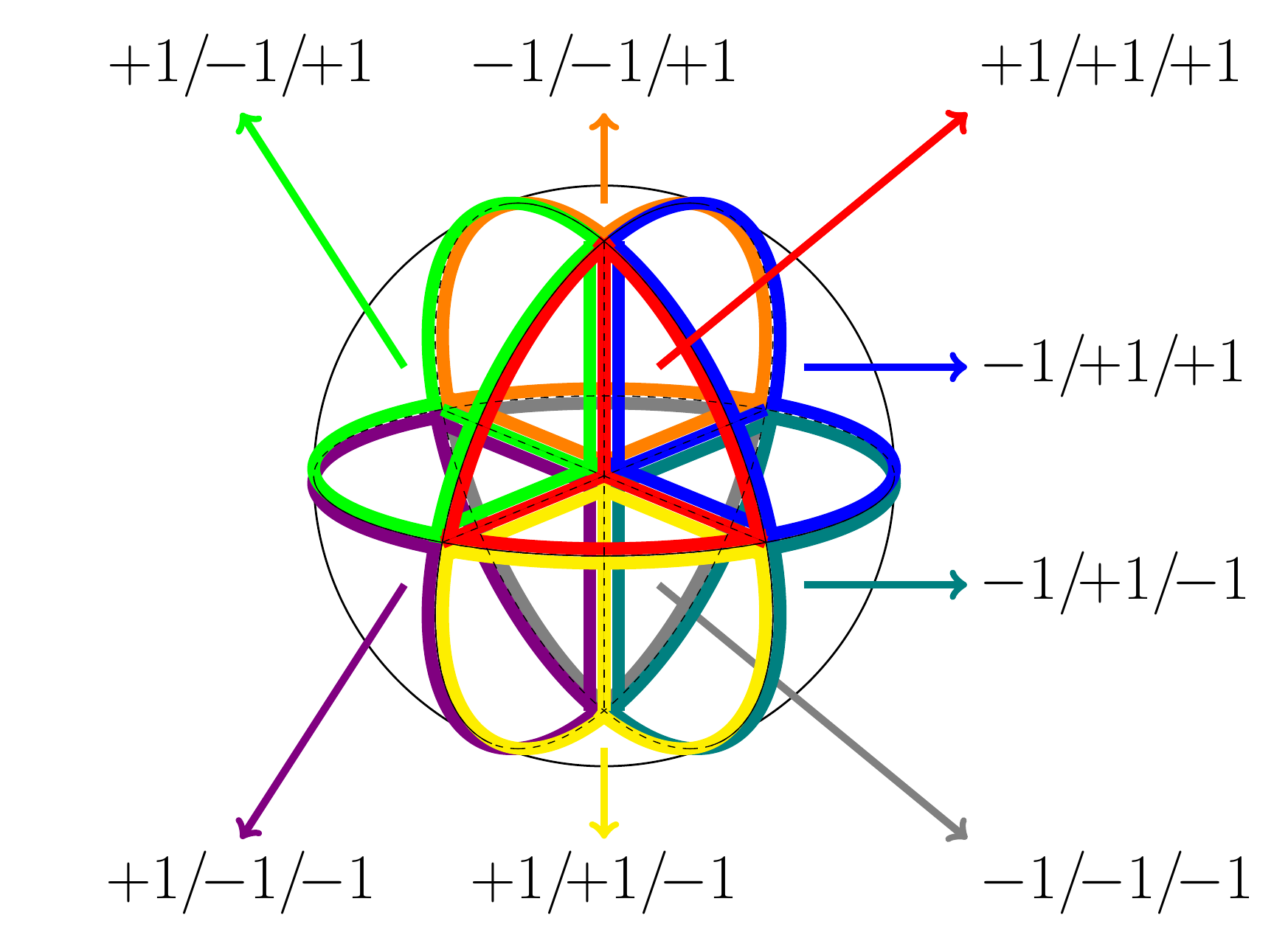}
\includegraphics[trim=15 0 15 0, scale=0.27]{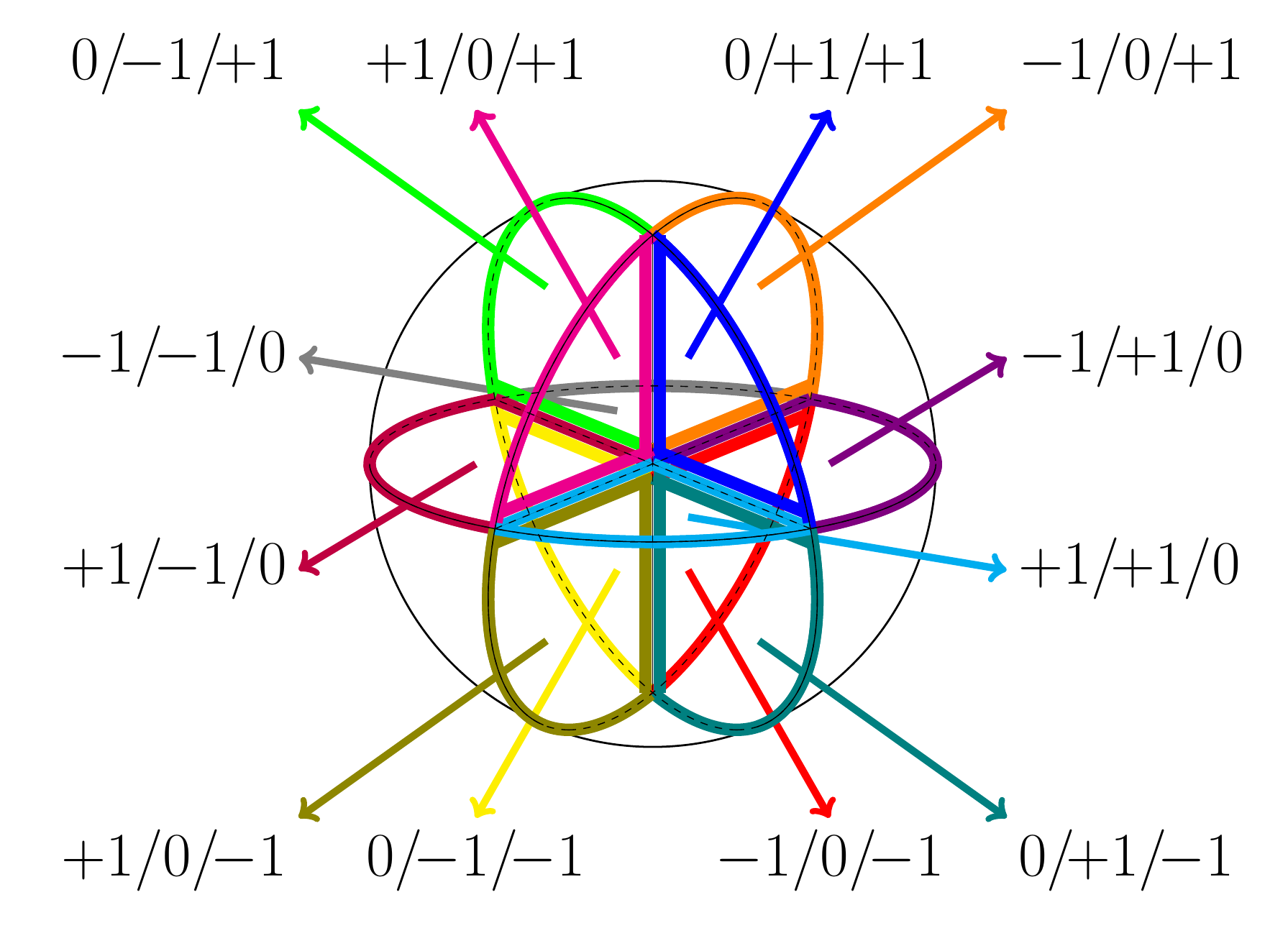}
\includegraphics[scale=0.27, trim=0 0 130 0]{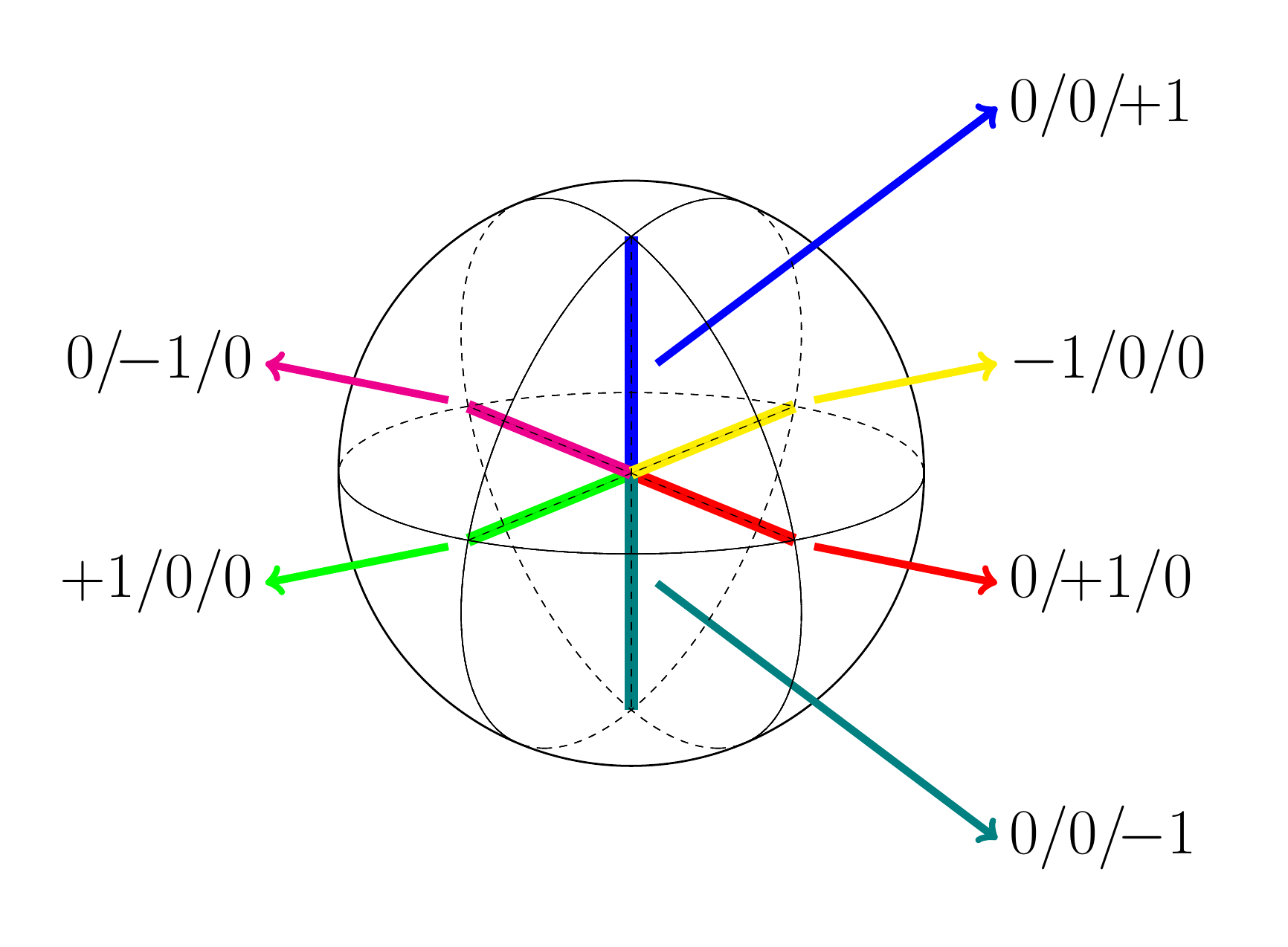}
\end{center}
\caption{Partial ordering of normal vectors. The numbers indicate the signs of normal vector components, which are used in the partial order function. The figures from left to right visualize different (ordered) sign combinations. Sets of sign combinations without zero values form volumes (left), sets with one zero value form planar areas (middle), and sets with two zero values form line sections (right). The 0/0/0 combination forms a point at the origin. \label{fig:signBalls}}
\end{figure}

There is an ambiguity in the representation of planes via normal vectors due to the two possible orientations of the normal vectors, which correspond to the same geometry. 
In general, there are two possibilities of including normal vector information into the model: (i) encoding always that triangle plane normal vector which always points towards or away from the corresponding particle, (ii) including positively and negatively oriented versions of the normal vector, and order them. We decided for option (ii) since pathological cases where the particle is in the same plane as the triangle are avoided and training is further stabilized.
In doing so, we have to deal with the fact that the network predictions should be invariant with respect to the orientation of the normal vectors. Therefore, we define a partial ordering which is able to sort the normal vectors with respect to their orientations. 
For a given normal vector $\Bn=\PAR{n_1,n_2,n_3} \in \dR^3$, we use the following partial order function
\begin{equation}
f_o\PAR{\Bn}=\sum_{i=1}^{3} 3^{i-1}\;\PAR{\sgn\PAR{n_i}+1}
\end{equation}
to retrieve the scalar values $f_o\PAR{\Bn}$ and $f_o\PAR{-\Bn}$ and sort the two vectors according to their corresponding mapped values. 
Different sign combinations of the normal vectors are shown in \cref{fig:signBalls}.
To test the performance of our approach, we conduct a toy experiment as well as a simulation experiment with different representations of normal vectors. We describe both experiments in \cref{sec:toyExp} and \cref{sec:simExp}.\\
It should be mentioned, that quite recently a new idea based on the ideas of \citet{Charles2017} and \citet{Zaheer2017} has been invented \citep{Lim2022} in a different context than ours and which was denoted as being a solution for "sign invariances". Their idea could also serve to be useful for resolving problems with orientation independent normal vector representations. While our approach works directly on the input vectors, their approach needs larger changes in the overall network architecture, but may on the other hand have advantages if the input itself would be the result of a DNN to get end-to-end differentiable architectures.

\subsection{Reflection Toy Example}
\label{sec:toyExp}

\begin{figure}[!htb]
	\centering
	\makebox[ \textwidth ]{
	\begin{annotationimage}[]{trim = 158 145 132 143, clip=True, height=150px}{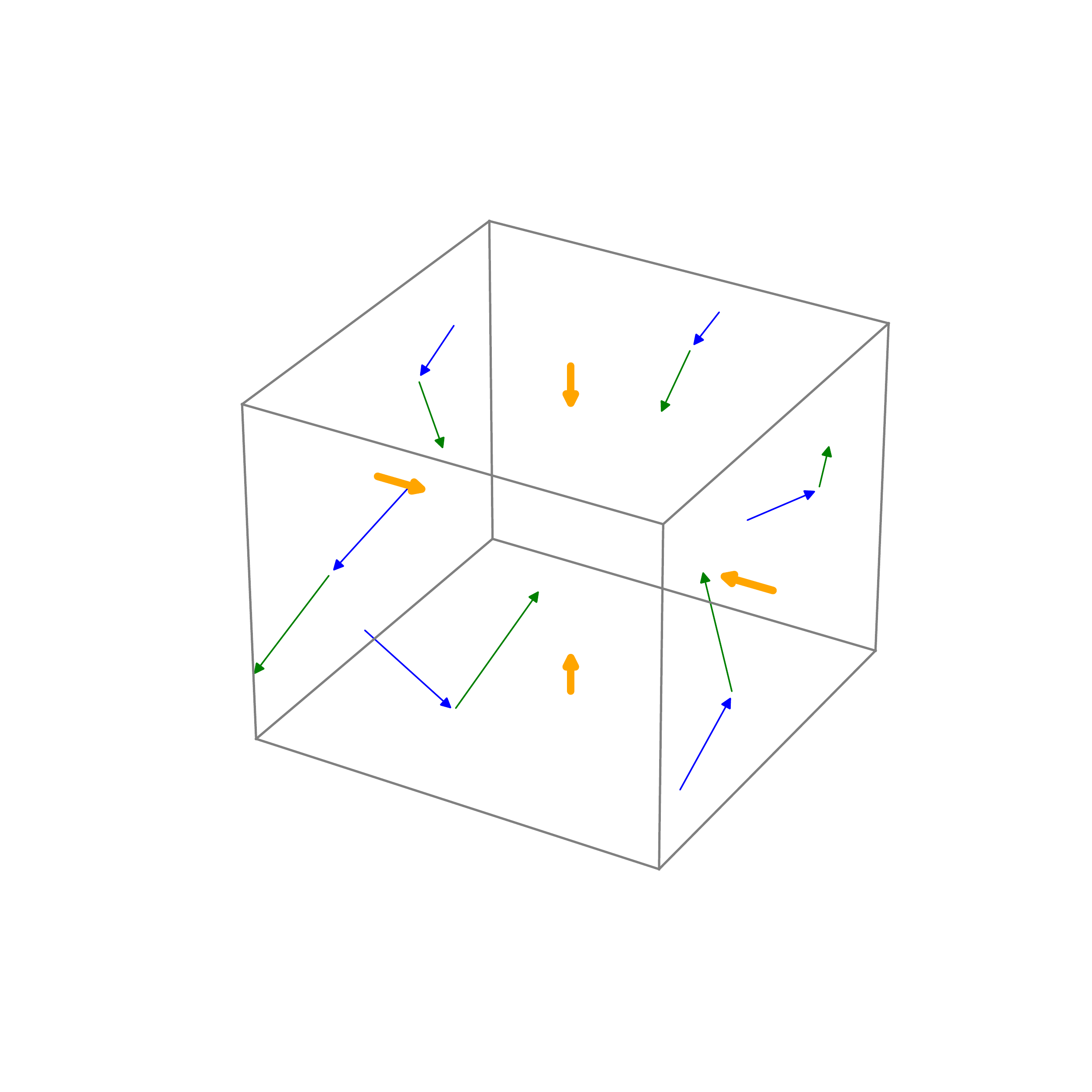}
	\draw[coordinate label = {ground truth: $\Bn$  at (0.4,1.08)}];
	\end{annotationimage}
	\hspace*{20px}
	\raisebox{40px}{
	\begin{overpic}[trim=395 333 285 271, clip=True, height=70px]{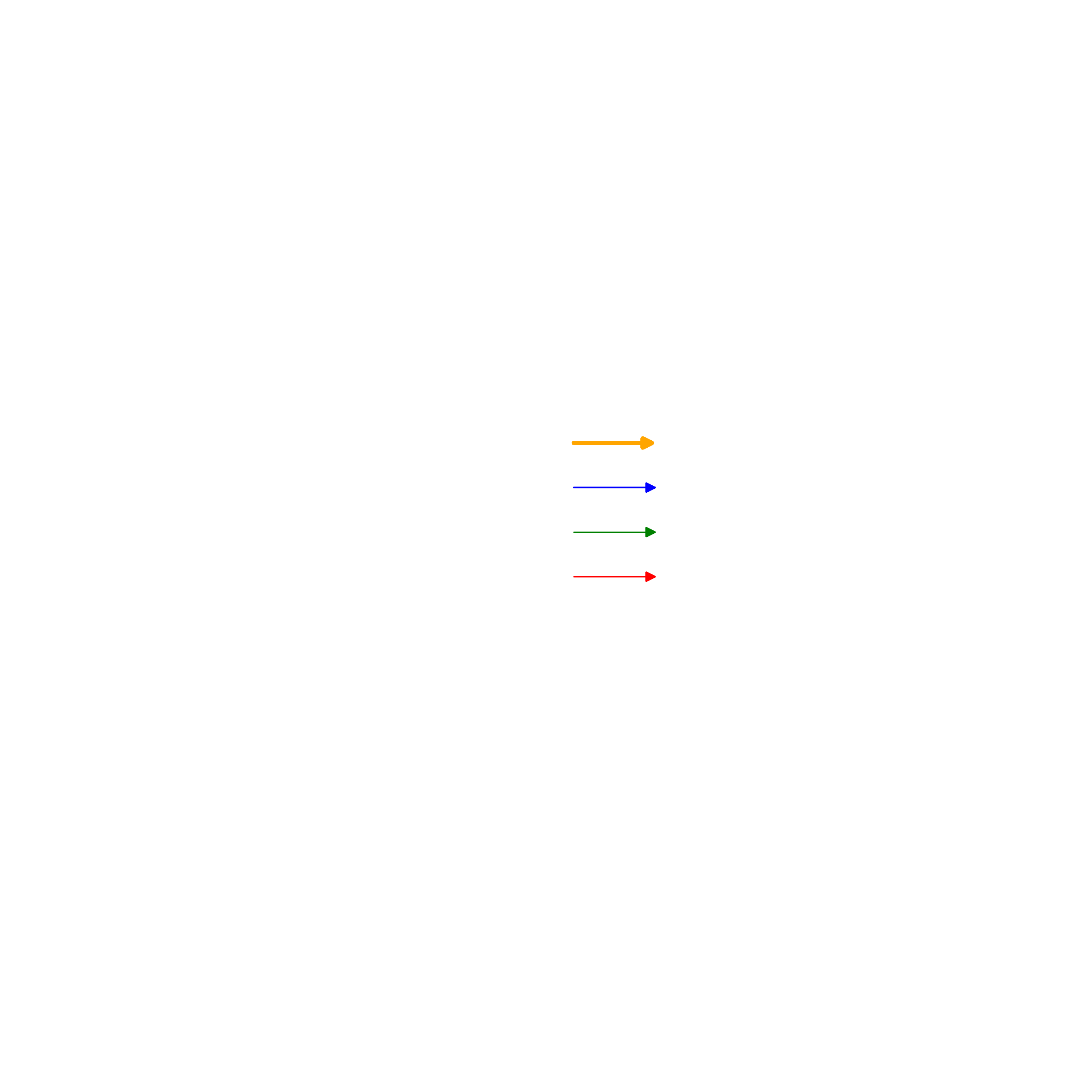}
	\put(43,77){wall normal vectors}
	\put(43,52){incident rays}
	\put(43,27){reflected rays}
	\put(43,2){predicted reflected rays}
	\end{overpic}
	}
	\hspace*{80px}
	
	}\\

    \vspace{0.5cm}
	\makebox[ \textwidth ]{
	\begin{annotationimage}[]{trim = 158 145 132 143, clip=True, height=150px}{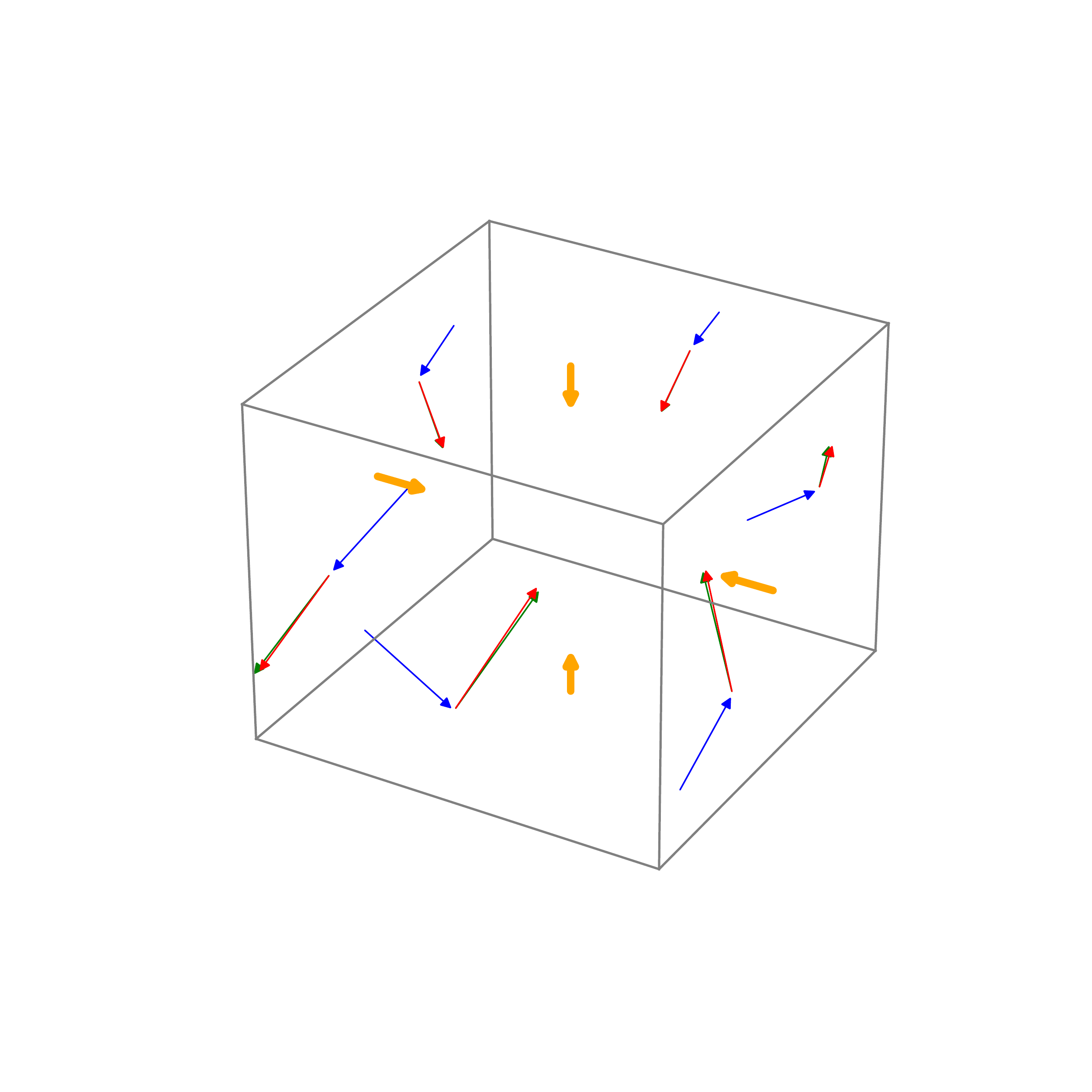}
	\draw[coordinate label = {R1: $\Bn$ at (0.4,1.08)}];
	\end{annotationimage}
	\begin{annotationimage}[]{trim = 158 145 132 143, clip=True, height=150px}{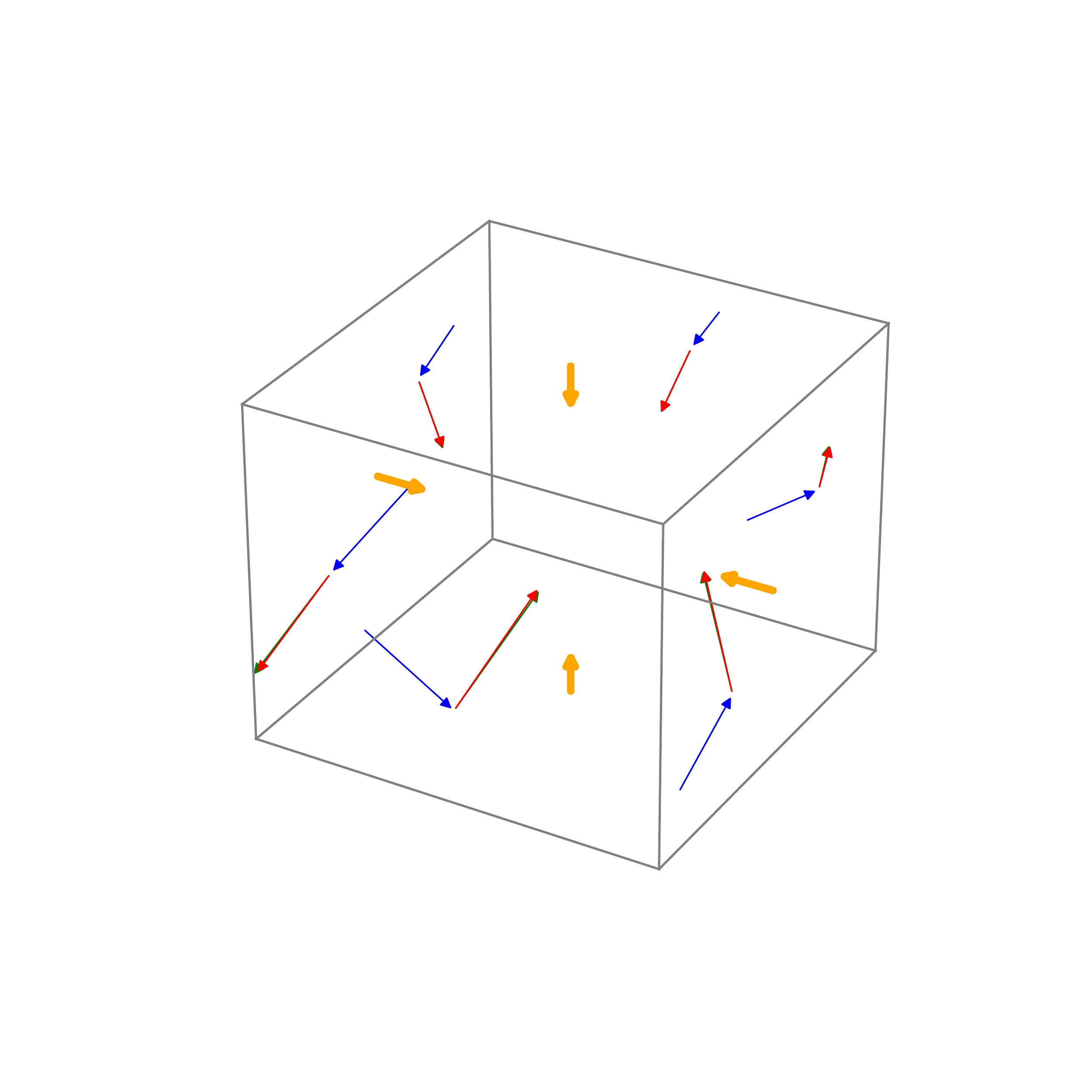}
	\draw[coordinate label = {R2: $\Bn, -\Bn$ at (0.4,1.08)}];
	\end{annotationimage}
	}
	\\
	\vspace{0.5cm}
	\makebox[ \textwidth ]{
	\begin{annotationimage}[]{trim = 158 145 132 143, clip=True, height=150px}{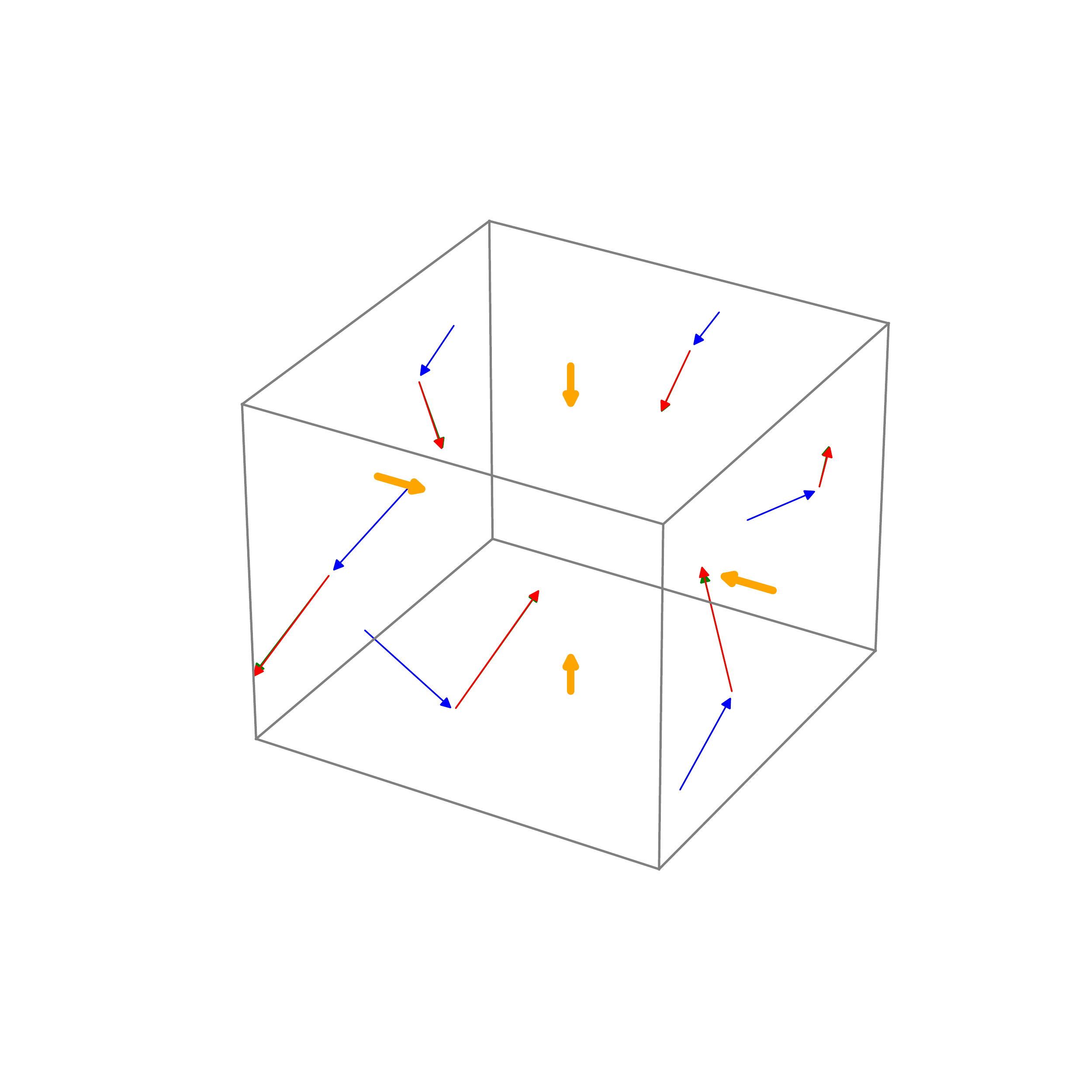}
	\draw[coordinate label = {R3: $\left\{
	\begin{array}{lc}
        \Bn & \text{if } \mathop{\mathbb{\mathrm{f_o}}}\PAR{\Bn} \leq \mathop{\mathbb{\mathrm{f_o}}}\PAR{-\Bn}\\
        -\Bn & \text{otherwise} 
    \end{array}
    \right.$ at (0.4,1.1)}];
	\end{annotationimage}
	\begin{annotationimage}[]{trim = 158 145 132 143, clip=True, height=150px}{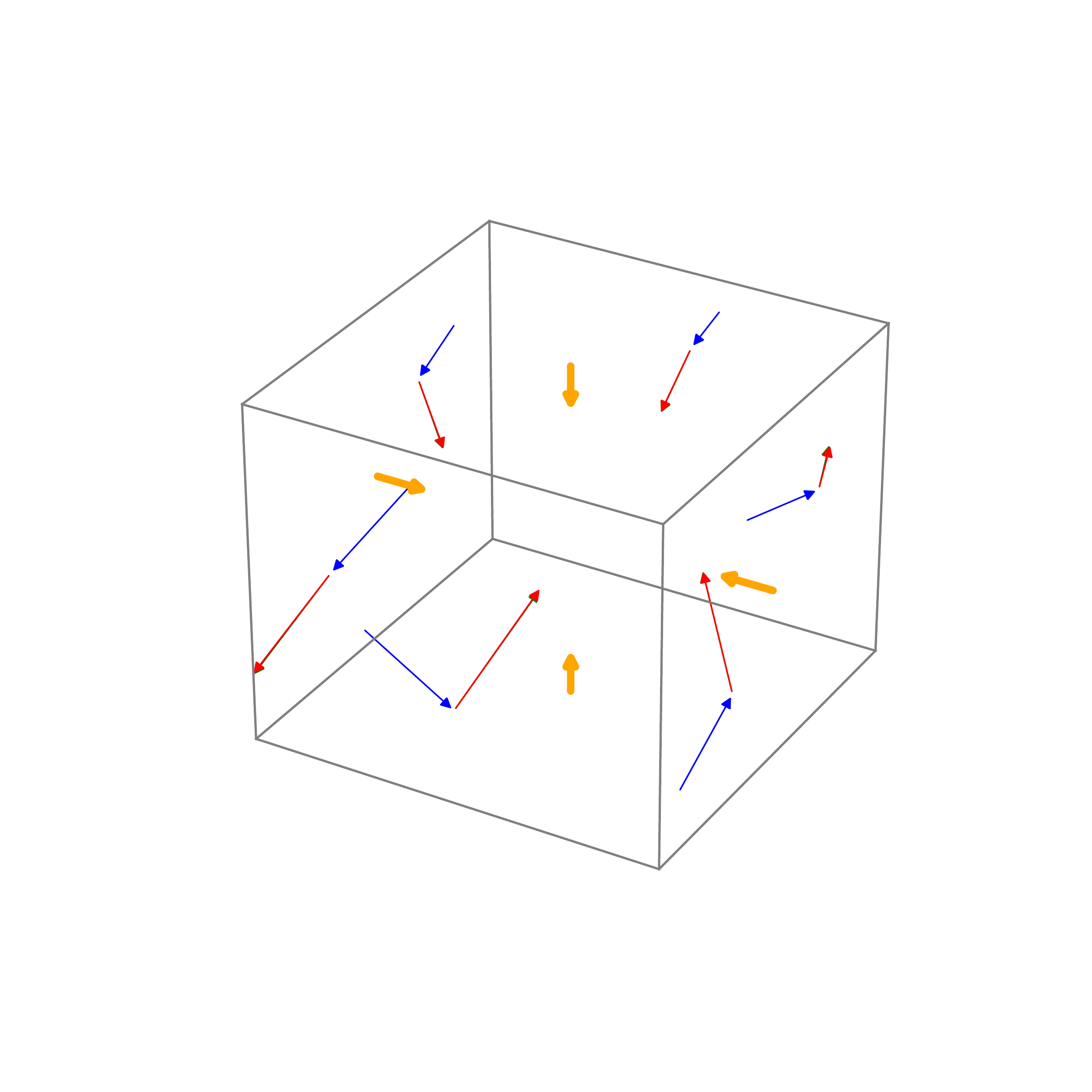}
	\draw[coordinate label = {R4: $\left\{
	\begin{array}{lc}
        \Bn,-\Bn & \text{if } \mathop{\mathbb{\mathrm{f_o}}}\PAR{\Bn} \leq \mathop{\mathbb{\mathrm{f_o}}}\PAR{-\Bn}\\
        -\Bn,\Bn & \text{otherwise} 
    \end{array}
    \right.$ at (0.4,1.1)}];
    \end{annotationimage}
	}
	\caption{Reflection of rays at four different walls (left, right, bottom, top). Wall normal vectors are visualized by bold orange arrows. The incident rays are visualized by blue arrows, reflected rays by green arrows, and neural network predictions by red arrows. Neural network predictions are based on wall representations \textbf{that are oriented the same way as in the training phase}. The caption above each plot indicates the wall input features used for training each of the networks. \label{fig:reflTrain} }
\end{figure}

\begin{figure}[!htb]
	\centering
	\makebox[ \textwidth ]{

	\begin{annotationimage}[]{trim = 158 145 132 143, clip=True, height=150px}{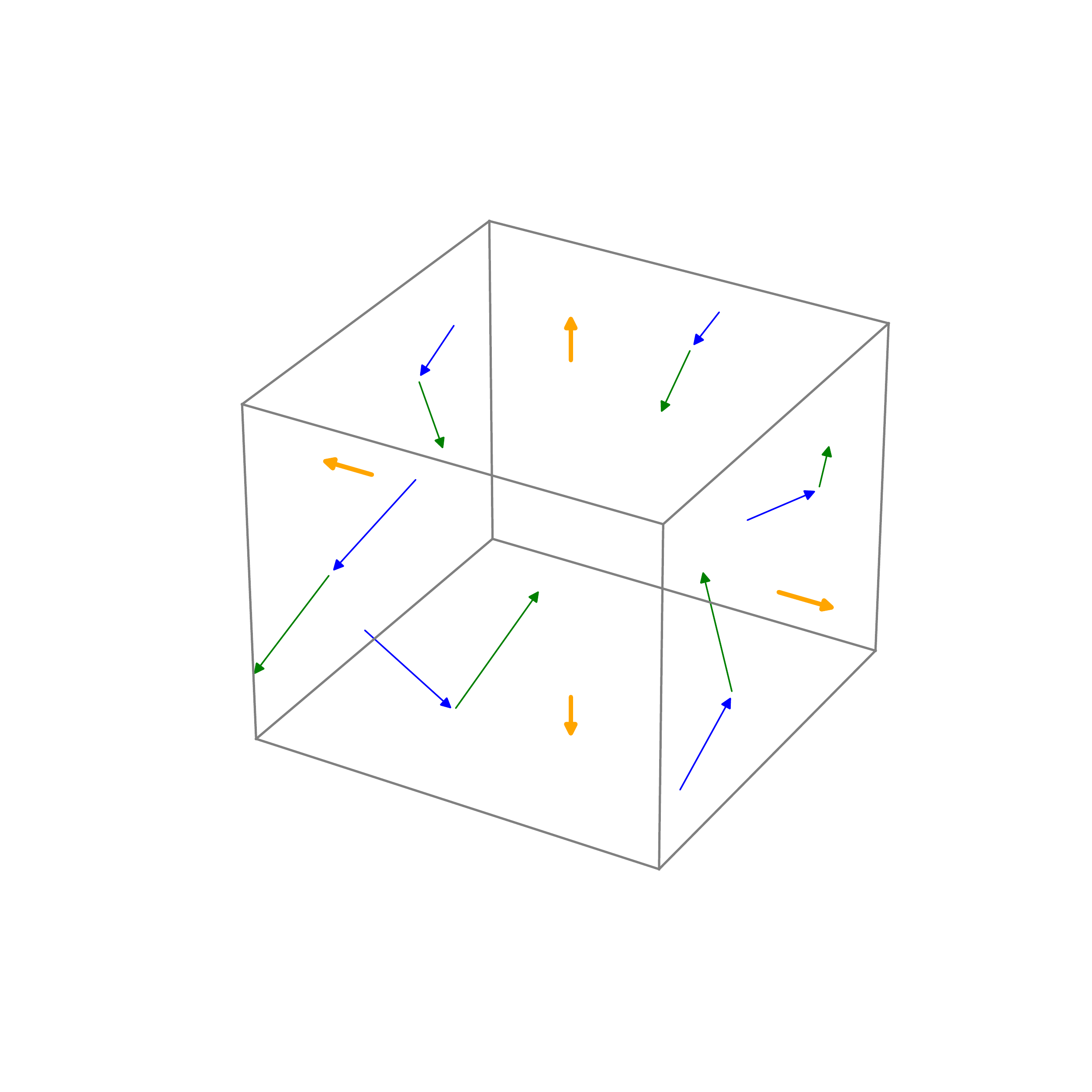}
	\draw[coordinate label = {ground truth: $-\Bn$  at (0.4,1.08)}];
	\end{annotationimage}
	\hspace*{20px}
	\raisebox{40px}{
	\begin{overpic}[trim=395 333 285 271, clip=True, height=70px]{figures/reflectionExp/legend.pdf}
	\put(43,77){wall normal vectors}
	\put(43,52){incident rays}
	\put(43,27){reflected rays}
	\put(43,2){predicted reflected rays}
	\end{overpic}
	}
	\hspace*{80px}
	}
	
    \vspace{0.5cm}
	\makebox[ \textwidth ]{
	\begin{annotationimage}[]{trim = 158 145 132 143, clip=True, height=150px}{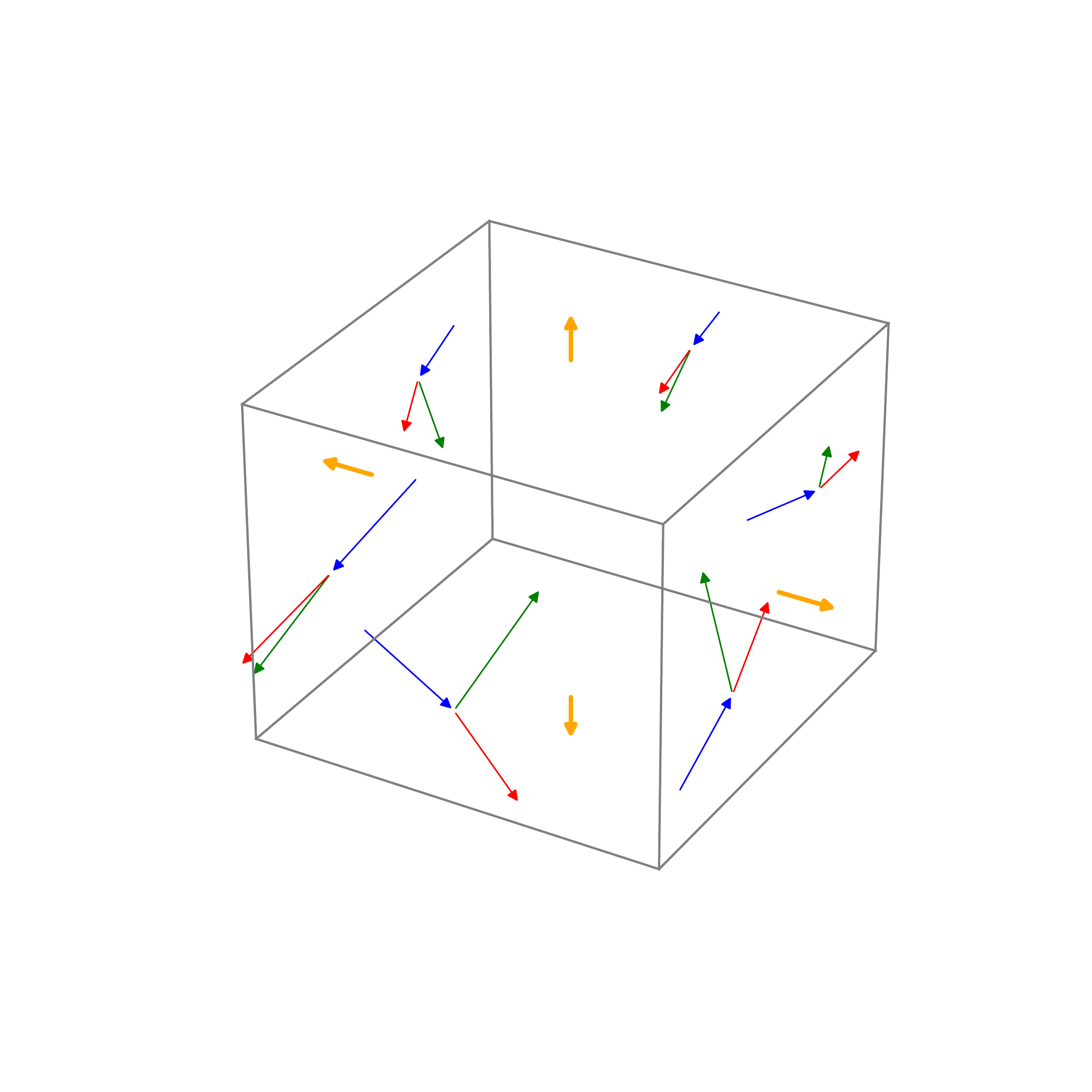}
	\draw[coordinate label = {R1: $-\Bn$ at (0.4,1.08)}];
	\end{annotationimage}
	\begin{annotationimage}[]{trim = 158 145 132 143, clip=True, height=150px}{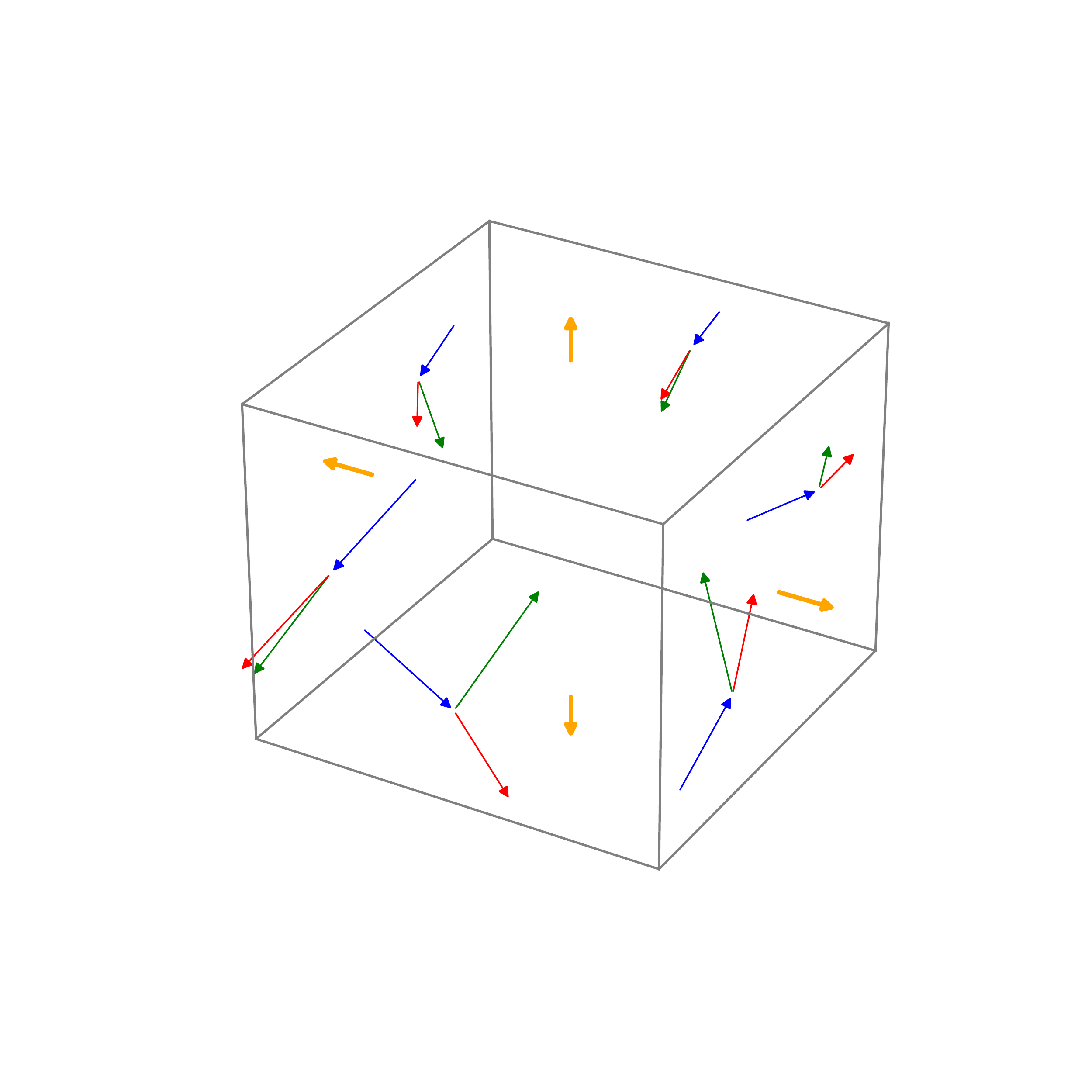}
	\draw[coordinate label = {R2: $-\Bn, \Bn$ at (0.4,1.08)}];
	\end{annotationimage}
	}\\
	\vspace{0.5cm}
	\makebox[ \textwidth ]{
	\begin{annotationimage}[]{trim = 158 145 132 143, clip=True, height=150px}{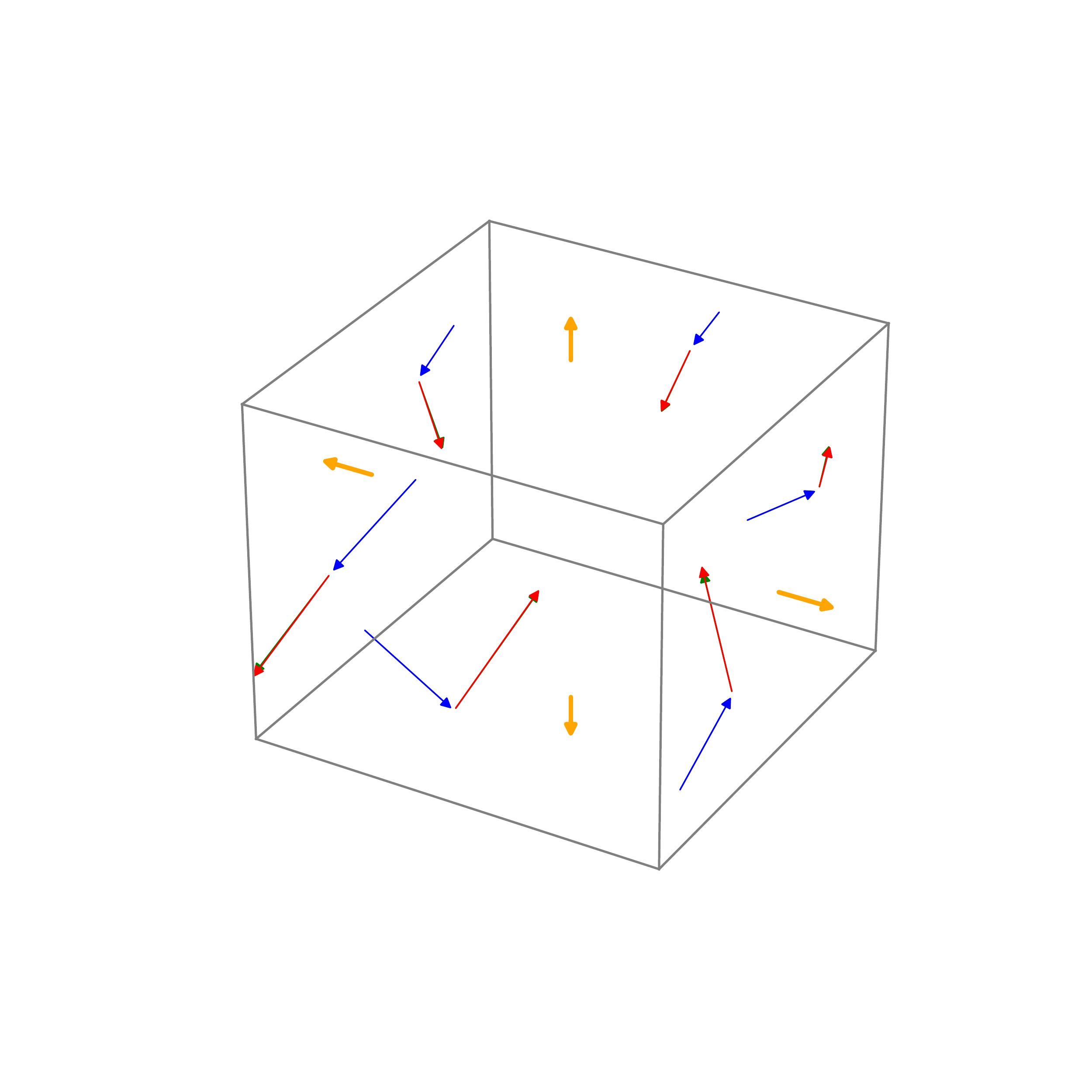}
	\draw[coordinate label = {R3: $\left\{
	\begin{array}{lc}
        -\Bn & \text{if } \mathop{\mathbb{\mathrm{f_o}}}\PAR{-\Bn} \leq \mathop{\mathbb{\mathrm{f_o}}}\PAR{\Bn}\\
        \Bn & \text{otherwise} 
    \end{array}
    \right.$ at (0.4,1.1)}];
	\end{annotationimage}
	\begin{annotationimage}[]{trim = 158 145 132 143, clip=True, height=150px}{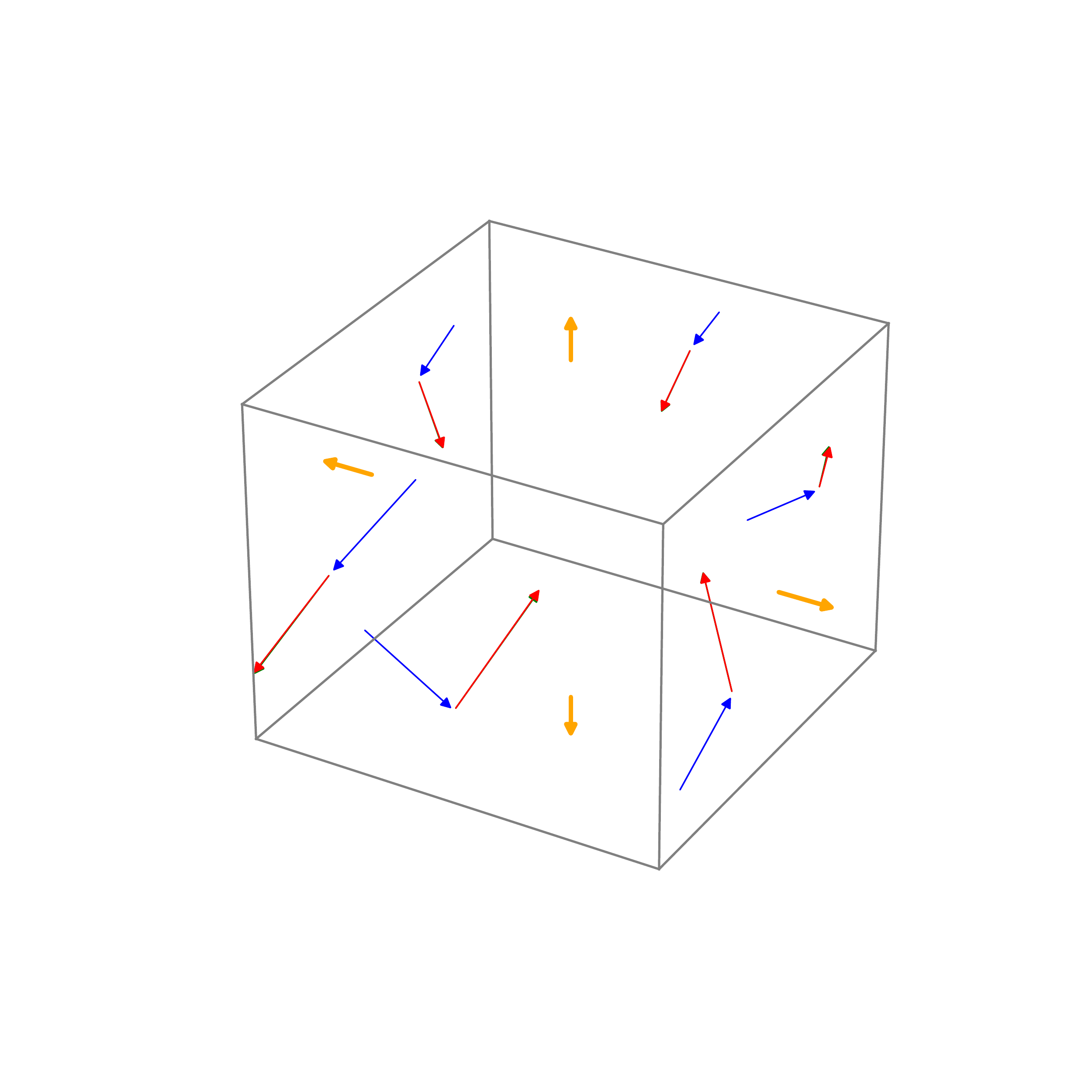}
	\draw[coordinate label = {R4: $\left\{
	\begin{array}{lc}
        -\Bn,\Bn & \text{if } \mathop{\mathbb{\mathrm{f_o}}}\PAR{-\Bn} \leq \mathop{\mathbb{\mathrm{f_o}}}\PAR{\Bn}\\
        \Bn,-\Bn & \text{otherwise} 
    \end{array}
    \right.$ at (0.4,1.1)}];
    \end{annotationimage}
	}
	\caption{Reflection of rays at four different walls (left, right, bottom, top). Wall normal vectors are visualized by bold orange arrows. The incident rays are visualized by blue arrows, reflected rays by green arrows, and neural network predictions by red arrows. Neural network predictions are based on wall representations 
	{\bf that are inversely oriented compared to the training phase}. The caption above each plot indicates the wall input features used for training each of the networks. \label{fig:reflTrainInv} }
\end{figure}

We conduct a toy experiment to showcase that a partial ordering of normal vectors is helpful for learning 3D simulations. In detail, 
we consider reflection ($Ref$) at a plane $\Bn$ as given by
\begin{align}
Ref_{\Bn}\left(\Bv\right) = \Bv - 2 \frac{\Bv \cdot \Bn}{\Bn \cdot \Bn} \Bn \ , 
\end{align}
and try to learn the reflection formula by a simple ReLU network, which takes the 3 components of $\Bn$ and $\Bv$ as input features and predicts the 3 components of $Ref_{\Bn}\left(\Bv\right)$. The training data consists of reflections at four fixed walls: the top, the bottom, the left, and, the right side of a simple cube. 
We use normal vectors of these walls, that point towards the inner of the cube. When evaluating the performance of the trained models, we observe decent predictions, if the orientation of the normal vectors describing the inclination of the walls was equal to the training data (see R1-R4 in \cref{fig:reflTrain}). However, for inverted normal vectors in the test set, only networks which take a partial ordering of the normal vectors into account predict the reflection correctly
(see R3, R4 in \cref{fig:reflTrainInv}).

\subsection{Simulation Experiment}
\label{sec:simExp}

We compare three different versions of how to include normal vector information for the hopper particle flow experiments:
\begin{itemize}
    \item not including normal vector information, and filling six node features up with zero entries instead (V1)
    \item including single normal vector orientation, which is given by the triangle corner point order of the mesh (V2)
    \item including both normal vector orientations (six features) (V3).
\end{itemize}
From an information perspective, it should be noted that (i) distance information (scalar distance and distance vectors) to the walls is  present in the edge features of the graph and (ii) in most cases the used normal vectors are oriented towards the outside of relevant border walls.
The different particle distribution trajectories obtained by the three versions are compared by computing the Earth Movers distances~\citep[][EMD]{Bonneel2011, Flamary2017pot} of predicted and simulated trajectories. We use Euclidean distances for the cost matrix, which we compute at time steps $2^0, 2^1, \cdots, 2^{16}$ for 5 training trajectories and 5 test trajectories. \Cref{tab:tbl_nvec} shows the means ($\mu$) and standard deviations ($\sigma$) of EMD values at different time steps and from 5 different training and test trajectories. A paired Wilcoxon test on the concatenated trajectories, shows that V3 significantly outperforms V1 (p-value 2.42e-04) and V2 (p-value 1.50e-03) on the test data. Interestingly, there is less significance on the training data, which might indicate that the usage of orientation-independent features to represent walls, helps to improve generalization performance, while it might not be that helpful for optimization purposes alone.

\begin{table}[ht]
\caption{Usage of different normal vector information in hopper particle flow experiment. The table summarizes means ($\mu$) and standard deviations ($\sigma$) of the EMD for the different versions and shows the results of a paired Wilcoxon test.}
\label{tab:tbl_nvec}
\begin{center}
\makebox[ \textwidth ]{
\begin{tabular}{ll|lll|lll}
\multicolumn{2}{c|}{\multirow{3}{*}{\bf \shortstack[c]{Version}}}
&\multicolumn{3}{c|}{\bf Train}
&\multicolumn{3}{c}{\bf Test}
\\
\multicolumn{2}{c|}{}
&\multicolumn{1}{c}{\multirow{2}{*}{\bf $\mu$}}  
&\multicolumn{1}{c}{\multirow{2}{*}{\bf $\sigma$}}
&\multicolumn{1}{c|}{\multirow{2}{*}{ \shortstack[c]{\small{p-value} \\ \tiny{Row $<$ V3}}}}
&\multicolumn{1}{c}{\multirow{2}{*}{\bf $\mu$}}
&\multicolumn{1}{c}{\multirow{2}{*}{\bf $\sigma$}}
&\multicolumn{1}{c}{\multirow{2}{*}{ \shortstack[c]{\small{p-value} \\ \tiny{Row $<$ V3}}}}
\\
\multicolumn{2}{c|}{}
&\multicolumn{3}{c|}{}
&\multicolumn{3}{c}{}
\\
\hline \multicolumn{2}{c|}{}
&\multicolumn{3}{c|}{}
&\multicolumn{3}{c}{}
\\

V1 & No normal vector     & 5.06e-05 & 1.17e-04 & 2.36e-02 & 6.80e-05 & 1.59e-04 & 2.42e-04\\
V2 & Single normal vector & 1.15e-04 & 3.84e-04 & 3.40e-03 & 1.21e-04 & 4.33e-04 & 1.50e-03\\
V3 & Both orientations    & 5.99e-05 & 1.77e-04 &          & 6.36e-05 & 2.06e-04 &         \\
\end{tabular}
}
\end{center}
\end{table}

\clearpage

\section{Experiments}\label{sec:app_experiments}

\subsection{Simulation details}

\paragraph{Hopper}
The initialization consists of two phases.
In a first step, particles are randomly inserted into a small cuboid which is positioned at a certain height above the closed hole of the hopper. This cuboid is continuously filled with particles during the initialization phase and afterwards particles freely move downwards (along the direction of gravity). In this way, the hopper is filled up to a certain height with 20,000 particles.
In a second phase, we cut out particles from the filled mass of particles.
We do this (i) by applying randomly selected functions and by (ii) randomly filtering out particles from the whole particle mass. The randomly selected functions are e.g. hyperplanes, where we only keep particles if they are at the same side of the hyperplane. The inserted particles have a radius of 0.002\,m.

\paragraph{Drum}
For initialization we assume that the direction of gravity is
different than the usual gravitation direction.
We insert particles at two random fixed regions within the drum. After the particles are inserted, they can move according to the gravitation direction during the initialization phase. In this way, we obtain different initial particle distributions within the drum. The inserted particles have a radius of 0.01\,m.

\subsection{Implementation details}

\paragraph{Graph Neural Network}
Raw inputs to our graph networks are initial particle positions and the particle positions from the 5 previous frames of the simulations. From these positions velocities are computed. Further inputs include the particle type and the coordinates of the triangle mesh of the respective time frame.
We use residual connections \citep{He2016} for both node and edge updates. For both updates, we use simple two-layer MLP networks, ReLUs \citep{Nair2010} after the first layer, and layer normalization \citep{Ba2016} without an additional activation after the second layer. For layer normalization we consider the $\epsilon$-parameter as a hyperparameter and set it to $1.0$. 
The networks for input embedding and read-out are similar to the message passing layers without layer normalization. The network weights are initialized similar to \citet{He2015Init}; for the input embeddings we assume an increased number of input neurons for \texttt{fan\_in}, where we consider the additional neurons as virtual copies of e.g. the wall indication feature in order to be able to upweight the influence of these features.
We use the mean-squared error as an objective and train with Adam optimization~\citep{Kingma2015}.
In order to facilitate learning, we provide as hyperparameter options not only $\NRM{ \hat{\Bx}_i-\hat{\Bx}_j}^2, \hat{\Bx}_i -\hat{\Bx}_j$ as features to the network, but also $\frac{1}{\NRM{ \hat{\Bx}_i-\hat{\Bx}_j}}, \frac{\hat{\Bx}_i -\hat{\Bx}_j}{\NRM{ \hat{\Bx}_i-\hat{\Bx}_j}^2}$ and $\frac{1}{\NRM{ \hat{\Bx}_i-\hat{\Bx}_j}^2}, \frac{\hat{\Bx}_i -\hat{\Bx}_j}{\NRM{ \hat{\Bx}_i-\hat{\Bx}_j}^3}$ reflecting the inverse distance law and the inverse-square law, which are present in many physical laws. We normalize input and target vectors and use a variant of Kahan summation \citep{kahan1965pracniques, klein2006generalized} in order to compute numerically stable statistics across particles of our dataset.

\paragraph{Hyperparameter Selection}
We keep 5 trajectories for each setting aside for validation. Criterions for hyperparameter selection are (i) that particles stay within the geometric object, and (ii) that the  ground truth trajectory is reproduced.

\subsection{Experimental Results For Cohesive Material}

\begin{figure}[!htb]
	\centering
	\begin{subfigure}[l]{0.45\textwidth}
	    \includegraphics[height=205px]{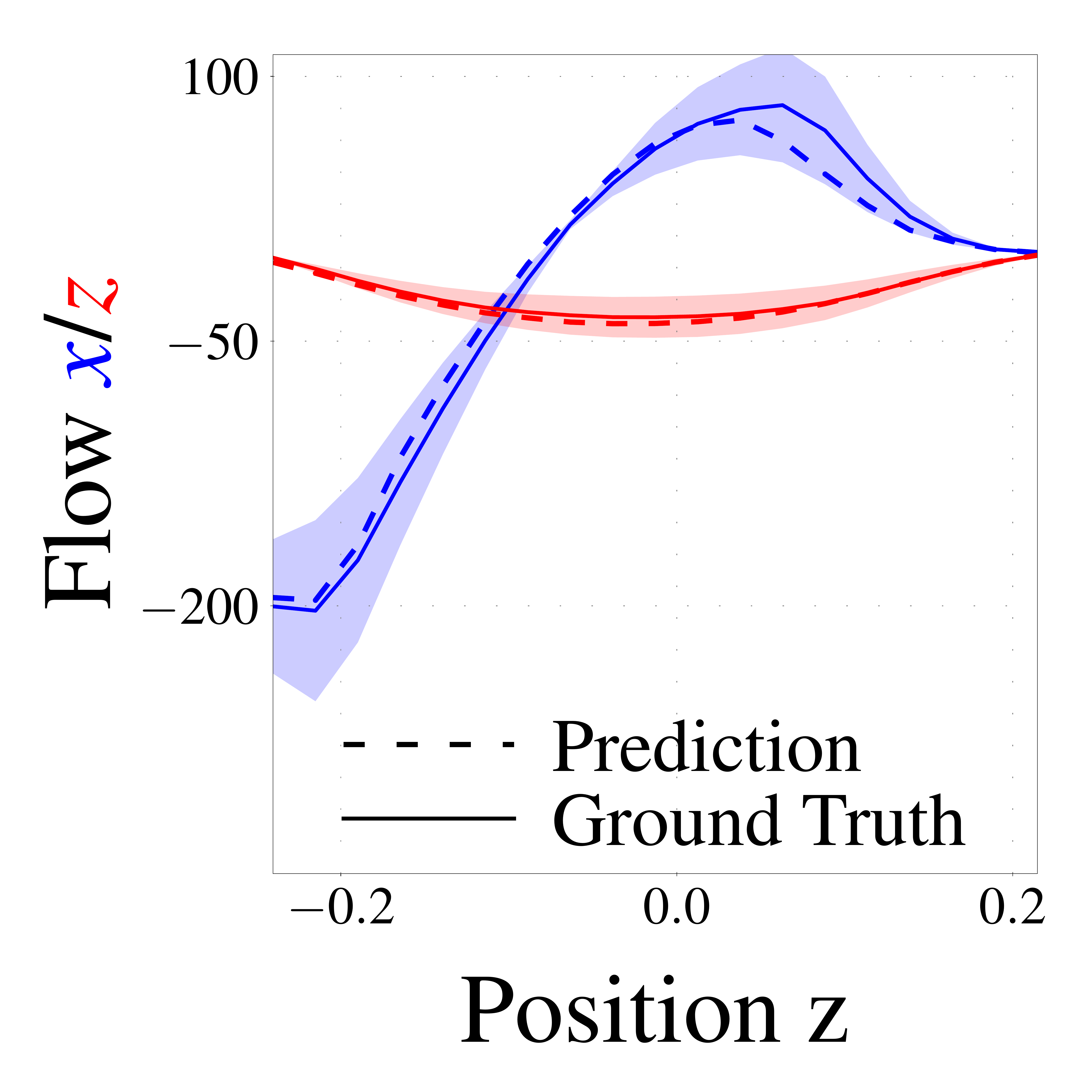}
	    \includegraphics[height=205px]{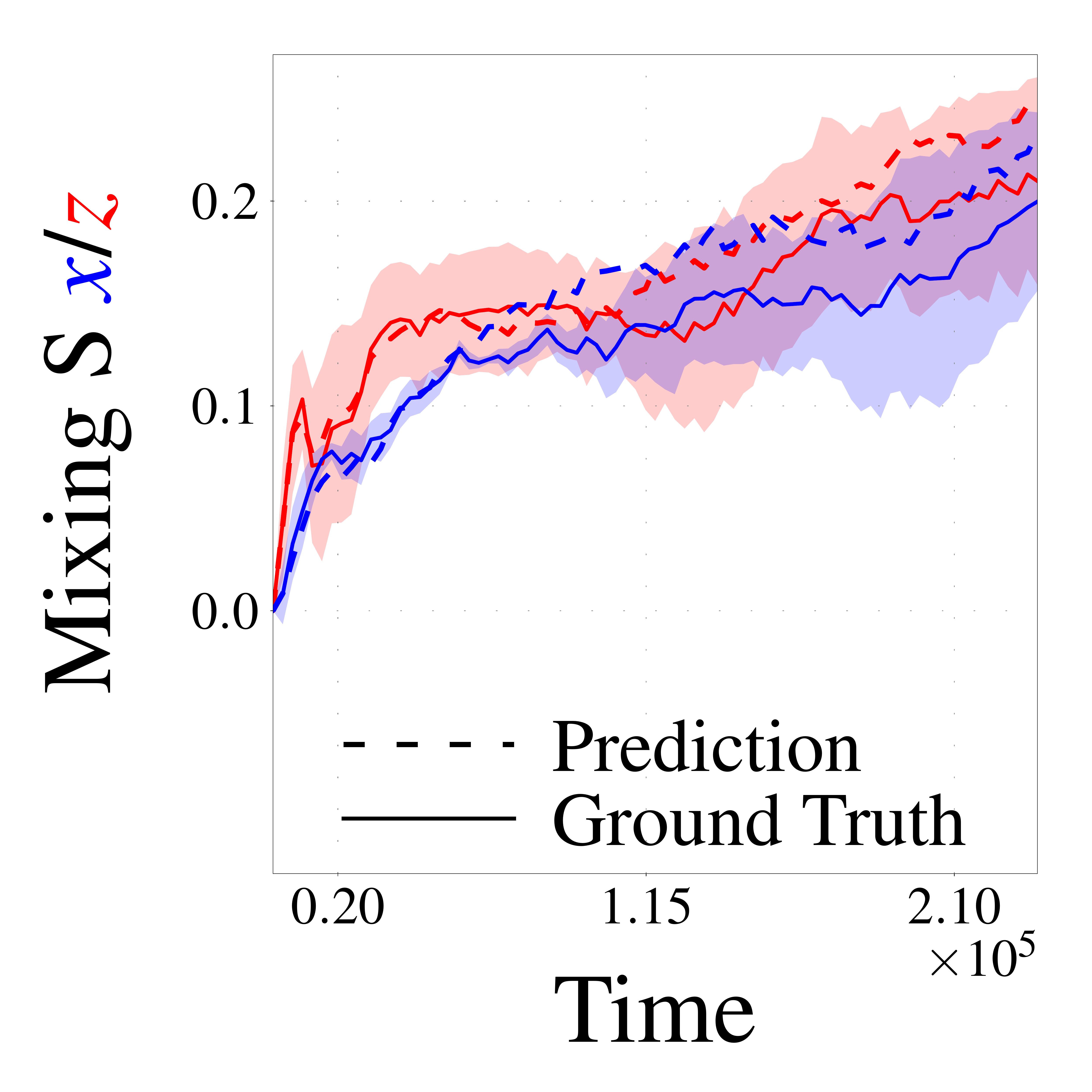}
	\end{subfigure}
    \hspace*{-0.cm}
	\begin{subfigure}[r]{0.45\textwidth}
	    \includegraphics[height=205px]{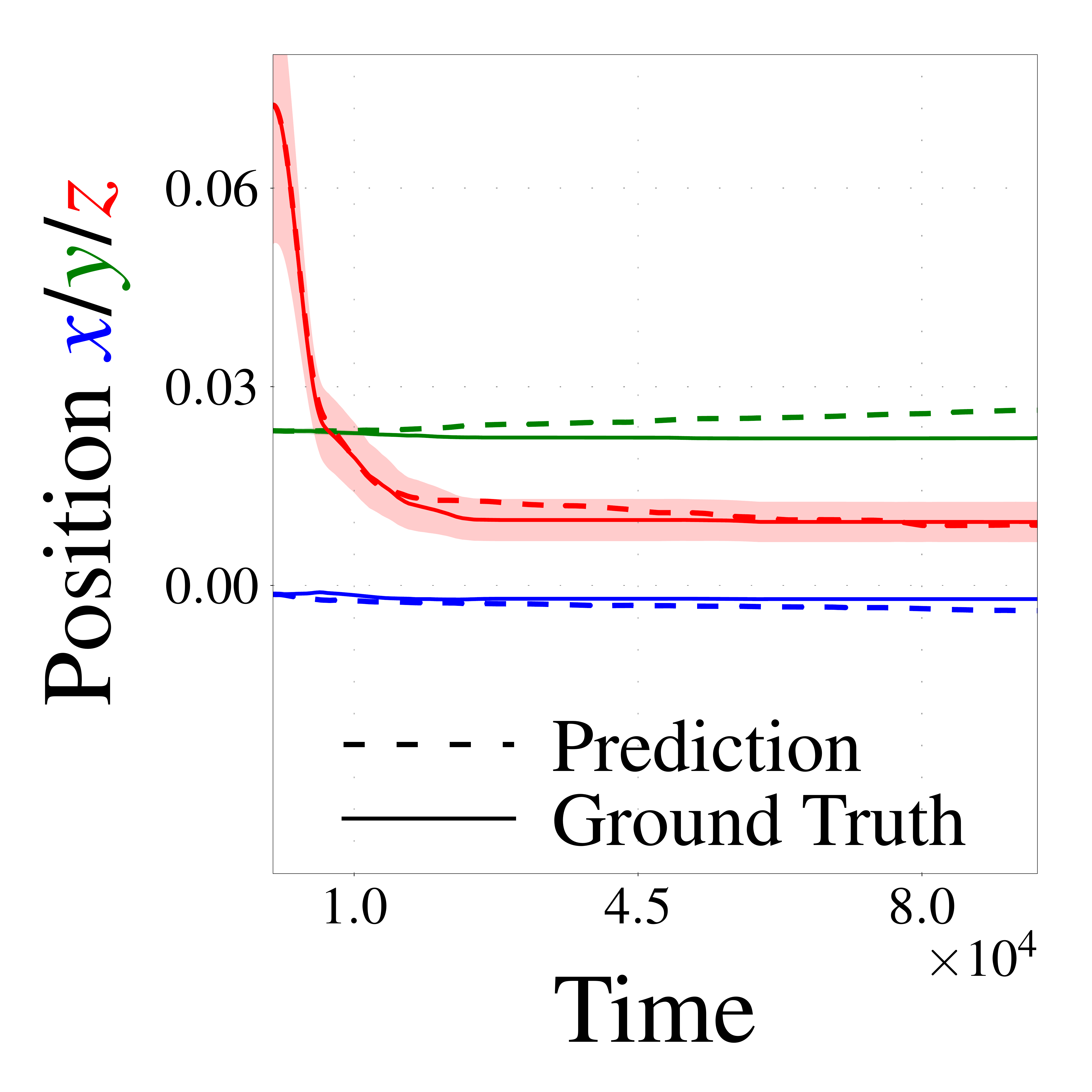}
	    \includegraphics[height=205px]{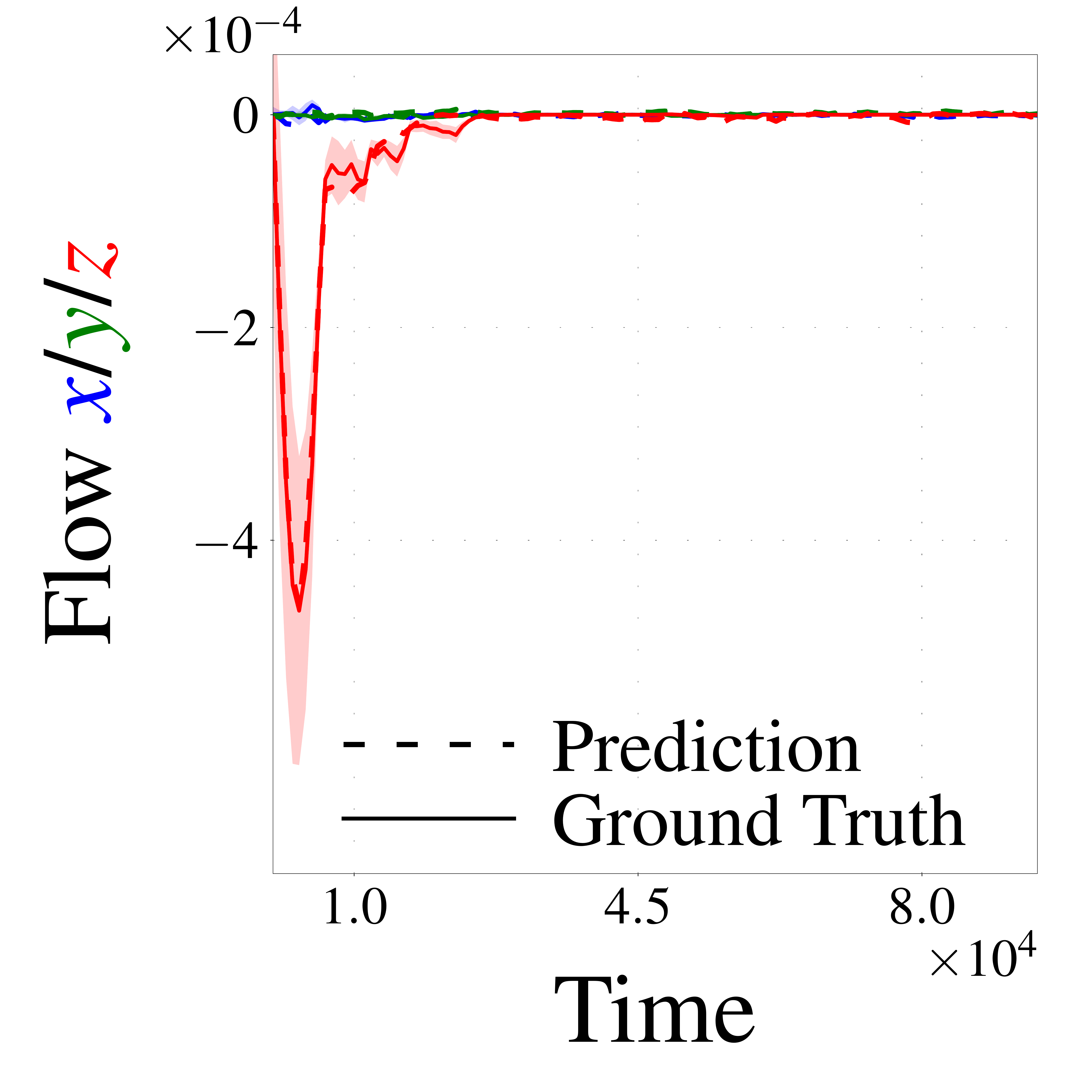}
	\end{subfigure}
\caption{Experimental results for cohesive granular meterial. Left, top: Time integrated particles flow in rotating drum in the x and z direction as a function of the position along the z axis. Left, bottom: Mixing entropies in the rotating drum as a function of time for particle class assignments according to the x (blue) and z (red) position. Right, top and bottom: Average particle position and particle flows for the hopper as a function of time.}	
	\label{fig:results_coh}
\end{figure}

In the following, experimental results for a cohesive material are shown. The results in the main paper are obtained for non-cohesive granular material, i.e. material with a cohesion energy density of $0$ J/$m^3$, which results in liquid-like behaviour. Increasing the cohesion energy density to $10^5$ J/$m^3$ corresponds to cohesive granular material, i.e. the particles have a strong tendency to clump together.
\Cref{fig:results_coh} shows the corresponding comparison of physical quantities for the cohesive granular material. Like in the non-cohesive case, the predictions for cohesive granular material are widely in agreement with the ground truth simulation.

\clearpage

\subsection{A word on hopper OOD experiments}

For hopper geometries (see e.g. \cref{fig:hopper_results}), OOD experiments are characterized by an increase of the side wall inclination angles and by an decrease of the radii of the outlet sizes. Especially due to the latter, we expect fewer particles to hit the ground for OOD architectures if particle-particle and particle-boundary interactions are correctly modeled.
In order to statistically test OOD trajectories against in-distribution trajectories,
we consider the proportion of particles which have traversed through the outlet of the hopper. We therefore create 15 in-distribution and 15 OOD trajectories for both cohesive and non-cohesive materials. We then apply a Mann-Whitney U test which assesses the proportion of in-distribution against the proportion of OOD particles traversing the outlet. The null hypotheses is that the same or a higher proportion of particles traverses the outlet for the case of an OOD trajectory compared to an in-distribution trajectory.

\begin{table*}[h]
  \centering
  \caption{Comparison of the proportion (mean $\mu$ and std $\sigma$) of particles beyond the outlet of the hopper}
  \footnotesize
  \begin{tabular}{l r r r r}
  \toprule
    & \multicolumn{2}{c}{ cohesive} & \multicolumn{2}{c}{non-cohesive}\\
    domain & $\mu$ & $\sigma$  & $\mu$ & $\sigma$ \\
    \midrule
    in-distribution & 0.34 & 0.09 & 0.89 & 0.03\\
    OOD & 0.14 & 0.11 & 0.73 & 0.15\\
    \bottomrule
  \end{tabular}
  \label{tab:}
\end{table*}

The Mann-Whitney U test shows that the predicted proportion values are significantly lower for OOD than for in-distribution trajectories (p-value  $\leq 1.4*10^{-4}$ for the cohesive material and p-value  $\leq 1.2*10^{-4}$ for the non-cohesive material). We remind the reader that this is expected due to the on average reduced outlet size in OOD geometries.
Furthermore, we compare the predicted proportion values of the cohesive and the non-cohesive model 
under the null hypothesis that the same or a higher proportion of particles traverses the outlet for the case of cohesive trajectories compared to non-cohesive trajectories. The applied Mann-Whitney U test yields a p-value $\leq 1.70*10^{-06}$ for the alternate hypothesis that the proportion is lower for the cohesive model, which is also in agreement with rational arguments (cohesive particles tend to clump together) and observations.

\subsection{Ablation studies}

For ablating BGNNs, we especially considered two design choices:
\begin{enumerate}
    \item Does sampling the triangularized boundaries also work?
    \item Is a unidirectional particle-wall interaction better suited to learn the corresponding particle-wall dynamics?
\end{enumerate}

\begin{figure}[!htb]
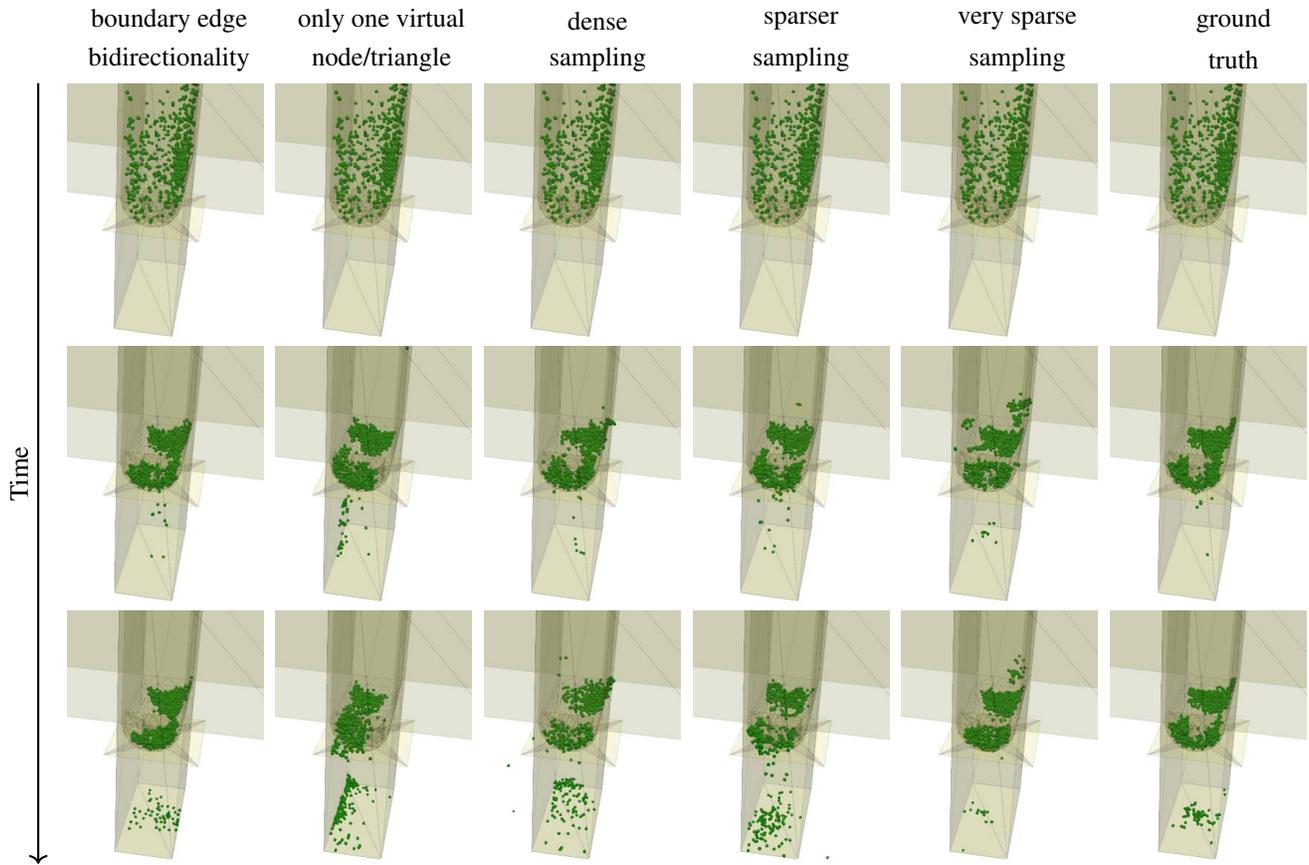


\centering

\begin{annotationimage}[]{trim = 5mm 207mm 93mm 0mm, clip=True, width=0.95\textwidth}{figures/results/ablation/hopper4}
\draw[coordinate label = {Cohesive Hopper Ablations at (0.5,1.18)}];

\draw[coordinate label = {boundary edge at (0.083,1.08)}];
\draw[coordinate label = { bidirectionality at (0.083,1.03)}];

\draw[coordinate label = { only one virtual at (0.25,1.08)}];
\draw[coordinate label = { node/triangle at (0.25,1.03)}];

\draw[coordinate label = { dense at (0.42,1.08)}];
\draw[coordinate label = { sampling at (0.42,1.03)}];

\draw[coordinate label = { sparser at (0.58,1.08)}];
\draw[coordinate label = { sampling at (0.58,1.03)}];

\draw[coordinate label = { very sparse at (0.75,1.08)}];
\draw[coordinate label = { sampling at (0.75,1.03)}];

\draw[coordinate label = {ground at (0.92,1.08)}];
\draw[coordinate label = {truth at (0.92,1.03)}];

    \draw [thick,->] (-0.02, 1.0)  -- node[anchor=south, rotate=90] {Time} (-0.02, 0);
\end{annotationimage}	
\caption{Ablation experiments for cohesive particles within hopper geometries. Different ablations from our default architecture (first five columns, see text) are compared to the ground truth (last column).  Particles are indicated by green spheres, triangular wall areas are yellow, the edges of these triangles are indicated by grey lines. 
}	
\label{fig:abl_hopper_coh}
\end{figure}

In \cref{fig:abl_hopper_coh,,fig:abl_hopper_ncoh,,fig:abl_drum_coh,,fig:abl_drum_ncoh} these points are assessed for cohesive and non-cohesive particles in hopper and rotating drum geometries, respectively. For the figures we kept hyperparameters consistent among the respective ablations (due to computational reasons). It should however be kept in mind, that this might not be optimal for the ablation, since individual hyperparameter optimization may lead to improved results. Nevertheless, we qualitatively tried to find an explanation what our trained ablation models might have learnt in order to avoid detected pitfalls. 

A consistent result across both geometries and particle types was that models with only sparse sampling may not capture particle dynamics correctly. For including bidirectional edges, the behaviour was so far inconclusive. 

In our ablation experiments with dense sampling, we tried to sample the triangle surfaces systematically, where the sampled points usually have distances that are in the range of particle diameters. For the sparser versions, we used multiples of the particle diameter (three, five). We found, that sampling might work well in the case of more homogeneous problems. For instance, in the case of cohesive particles, that tend to stick together, the model has learnt to avoid that a cluster of particles can move through boundary walls (see sparse sampling in \cref{fig:abl_hopper_coh}). Dense sampling might also work well for simple, homogeneous geometric settings like a drum. Although some particles for dense sampling are wrongly predicted to be at the top of the drum mesh in \cref{fig:abl_drum_coh}, which may have occurred due too non-optimal hyperparameter settings, dense sampling has captured the basic dynamics well. This is not any more the case for sparser settings, where especially for the non-cohesive case, particles may freely move through the walls (see, e.g., empty drum in \cref{fig:abl_drum_ncoh}).
It is interesting, that the model relying on dense sampling in \cref{fig:abl_hopper_ncoh} did not capture the moving dynamics at the ground very well. A reason for this could be, that the limited set of training samples, did not contain geometrically similar interactions between real and sampled virtual particles.

It should be mentioned wrt. to the experiment on bidirectional particle - wall edges, that our idea for unidirectional edges was, that a symmetry break in message passing might be helpful in learning particle - wall interactions.
The ablation experiments in this section however showed that also architectures without the symmetry break may in principle be able to learn reasonable behavior. 
Related to node connectivity, we also checked, whether it would be better to insert one respective virtual triangle node per neighboring real particle or whether we could use one virtual super-node per triangle for all edges instead. In both cases, the edges carry information on the minimum triangle-point distances. There may be subtle differences, which were likely induced by different weightings of the nodes.

\begin{figure}[!htb]

\centering

\begin{annotationimage}[]{trim = 5mm 207mm 93mm 0mm, clip=True, width=0.95\textwidth}{figures/results/ablation/hopper5}
\draw[coordinate label = {Non-Cohesive Hopper Ablations at (0.5,1.18)}];

\draw[coordinate label = {boundary edge at (0.083,1.08)}];
\draw[coordinate label = { bidirectionality at (0.083,1.03)}];

\draw[coordinate label = { only one virtual at (0.25,1.08)}];
\draw[coordinate label = { node/triangle at (0.25,1.03)}];

\draw[coordinate label = { dense at (0.42,1.08)}];
\draw[coordinate label = { sampling at (0.42,1.03)}];

\draw[coordinate label = { sparser at (0.58,1.08)}];
\draw[coordinate label = { sampling at (0.58,1.03)}];

\draw[coordinate label = { very sparse at (0.75,1.08)}];
\draw[coordinate label = { sampling at (0.75,1.03)}];

\draw[coordinate label = {ground at (0.92,1.08)}];
\draw[coordinate label = {truth at (0.92,1.03)}];

    \draw [thick,->] (-0.02, 1.0)  -- node[anchor=south, rotate=90] {Time} (-0.02, 0);
\end{annotationimage}	

\caption{Ablation experiments for non-cohesive particles within hopper geometries.
Different ablations from our default architecture (first five columns, see text) are compared to the ground truth (last column).  Particles are indicated by green spheres, triangular wall areas are yellow, the edges of these triangles are indicated by grey lines. 
}	
\label{fig:abl_hopper_ncoh}
\end{figure}

\begin{figure}[!htb]

\centering

\begin{annotationimage}[]{trim = 5mm 222mm 94mm 0mm, clip=True, width=0.95\textwidth}{figures/results/ablation/drum9}
\draw[coordinate label = {Cohesive Rotating Drum ablations at (0.5,1.20)}];

\draw[coordinate label = {boundary edge at (0.083,1.09)}];
\draw[coordinate label = { bidirectionality at (0.083,1.03)}];

\draw[coordinate label = { only one virtual at (0.25,1.09)}];
\draw[coordinate label = { node/triangle at (0.25,1.03)}];

\draw[coordinate label = { dense at (0.42,1.09)}];
\draw[coordinate label = { sampling at (0.42,1.03)}];

\draw[coordinate label = { sparser at (0.58,1.09)}];
\draw[coordinate label = { sampling at (0.58,1.03)}];

\draw[coordinate label = { very sparse at (0.75,1.09)}];
\draw[coordinate label = { sampling at (0.75,1.03)}];

\draw[coordinate label = {ground at (0.92,1.09)}];
\draw[coordinate label = {truth at (0.92,1.03)}];

    \draw [thick,->] (-0.02, 1.0)  -- node[anchor=south, rotate=90] {Time} (-0.02, 0);
\end{annotationimage}
\caption{Ablation experiments for cohesive particles within rotating drum geometries. Different ablations from our default architecture (first five columns, see text) are compared to the ground truth (last column).  Particles are indicated by green spheres, triangular wall areas are yellow, the edges of these triangles are indicated by grey lines. The circular arrow indicates the rotation direction of the drum.
}	
\label{fig:abl_drum_coh}
\end{figure}

\begin{figure}[!htb]
\centering
\begin{annotationimage}[]{trim = 5mm 222mm 94mm 0mm, clip=True, width=0.95\textwidth}{figures/results/ablation/drum10}
\draw[coordinate label = {Non-Cohesive Rotating Drum ablations at (0.5,1.20)}];

\draw[coordinate label = {boundary edge at (0.083,1.09)}];
\draw[coordinate label = { bidirectionality at (0.083,1.03)}];

\draw[coordinate label = { only one virtual at (0.25,1.09)}];
\draw[coordinate label = { node/triangle at (0.25,1.03)}];

\draw[coordinate label = { dense at (0.42,1.09)}];
\draw[coordinate label = { sampling at (0.42,1.03)}];

\draw[coordinate label = { sparser at (0.58,1.09)}];
\draw[coordinate label = { sampling at (0.58,1.03)}];

\draw[coordinate label = { very sparse at (0.75,1.09)}];
\draw[coordinate label = { sampling at (0.75,1.03)}];

\draw[coordinate label = {ground at (0.92,1.09)}];
\draw[coordinate label = {truth at (0.92,1.03)}];

    \draw [thick,->] (-0.02, 1.0)  -- node[anchor=south, rotate=90] {Time} (-0.02, 0);
\end{annotationimage}
\caption{ Ablation experiments for non-cohesive particles withing rotating drum geometries. Different ablations from our default architecture (first five columns, see text) are compared to the ground truth (last column).  Particles are indicated by green spheres, triangular wall areas are yellow, the edges of these triangles are indicated by grey lines. The circular arrow indicates the rotation direction of the drum.
}	
\label{fig:abl_drum_ncoh}
\end{figure}

\clearpage
\renewcommand{\bibsection}{\section*{TApp. References}}
\bibliographystyle {aaai23}
\bibliography{refsApp}

\end{document}